\documentclass[runningheads]{llncs}

\usepackage{eccv}

\usepackage{eccvabbrv}
\usepackage{multirow}
\usepackage{graphicx}
\usepackage{booktabs}
\usepackage{makecell}
\usepackage[accsupp]{axessibility}  

\usepackage{amssymb}
\usepackage{xcolor}
\usepackage{tcolorbox}

\usepackage{pifont}   

\usepackage{hyperref}

\usepackage{orcidlink}

\begin{document}

\title{Rolling Shutter Camera Self-Calibration} 

\titlerunning{Rolling Shutter Camera Self-Calibration}

\author{Yongcong Zhang\inst{1,2,*}\orcidlink{0009-0009-1254-6549} \and 
Navid Rabbani\inst{1,*}\orcidlink{0009-0006-2409-4824} \and
Bangyan Liao\inst{3,*}\orcidlink{0009-0007-7739-4879} \and \\
Chengbo Wang\inst{4}\orcidlink{0009-0009-1385-1633} \and 
Yizhen Lao\inst{4}\orcidlink{0000-0002-6284-1724} \and
Adrien Bartoli\inst{1,2,\dagger}\orcidlink{0000-0003-3545-7329}
}   

\authorrunning{Y. Zhang, N. Rabbani et al.}


\institute{
\textsuperscript{1}{DIA2M, DRCI, CHU Clermont-Ferrand, France} \\
\textsuperscript{2}SURGAR, Surgical Augmented Reality, Clermont-Ferrand, France \\
\textsuperscript{3}School of Engineering, Westlake University, China \\ 
\textsuperscript{4}School of Design, Hunan University, China
}
\maketitle
\renewcommand{\thefootnote}{\relax}\footnotetext{
\textsuperscript{$\star$} Equal contribution\\  
\inst{\dagger} Corresponding author: \email{adrien.bartoli@gmail.com} \\
Project page: \href{https://github.com/Yongcong-Zhang/RSSC}{https://github.com/Yongcong-Zhang/RSSC}}
\renewcommand{\thefootnote}{\arabic{footnote}}
\setcounter{footnote}{0}


\begin{abstract}
Rolling shutter (RS) cameras are widely used in consumer devices, but their row-wise exposure causes distortions under motion, making geometric 3D vision problems dependent on both camera intrinsics and readout time ratio. Existing RS calibration methods rely on calibration targets or specialised hardware, limiting their use in unconstrained settings. We present the first self-calibration method for RS cameras that directly estimates camera intrinsics and the readout time ratio from image sequences, without requiring calibration targets. The method is implemented as a self-calibrating bundle adjustment (BA), which critically depends on the RS imaging model. We combine two known complementary models. The first formulates RS imaging as continuous-time trajectory estimation under a row-wise pose representation. The second interprets RS images as temporally distorted global shutter (GS) images and requires to estimate correction fields. The combination is non-trivial and results in a unified dual-projection model, in which each 3D point is simultaneously constrained at both row-dependent and reference timestamps along a shared continuous trajectory, enforcing stronger geometric and temporal consistency. Extensive simulations analyse the applicability of several implementations under varying conditions, and real data experiments demonstrate the accuracy, robustness, and practical effectiveness of the proposed approach.

\end{abstract}

\section{Introduction}
\label{sec:Introduction}
RS cameras are widely used in modern consumer imaging devices due to their low cost, low power consumption, and high frame rate~\cite{janesick2009fundamental, qu2023fast}. Unlike GS cameras, which capture the entire image at a single instant, RS cameras expose and read out sequentially in a row-wise manner. As a consequence, different image rows correspond to slightly different time points. Under camera or scene motion, this sequential readout introduces characteristic geometric distortions such as skew, wobble, and the well-known jello effect~\cite{meingast2005geometric}.
These RS distortions significantly complicate many fundamental vision tasks, including camera calibration~\cite{oth2013rolling,huai2022continuous,ito2017self,huai2023automated}, absolute and relative pose estimation~\cite{albl2019rolling,dai2016rolling}, as well as BA~\cite{hedborg2012rolling,liao2023revisiting} in SfM~\cite{albl2016degeneracies,lao2018robustified} and SLAM~\cite{schubert2018direct,kim2016direct} systems. 

The availability of accurate camera parameters is essential for geometric 3D vision problem, such as 3D reconstruction~\cite{tomasi1992shape, hartley2003multiple}. 
Compared with the GS model, RS imaging requires not only accurate intrinsic calibration but also an explicit modeling of the temporal readout process, depending on the readout time ratio $\gamma$ (Fig.~\ref{fig: RS exposure}(a)).
This extra parameter is defined as the ratio between the exposure readout time and the inter-frame interval~\cite{liang2008analysis, zhuang2017rolling, qu2023fast, qu2023towards}; it strongly controls the magnitude of the RS distortion and is widely adopted in RS image rectification~\cite{liang2008analysis, zhuang2017rolling, qu2023fast, qu2023towards,albl2020two} and camera pose estimation~\cite{zhuang2017rolling, zhang2024ego}. However, unlike standard intrinsic parameters, $\gamma$ cannot be obtained from GS calibration procedures.

\begin{figure}[t]
\centering
\includegraphics[width=1\columnwidth]{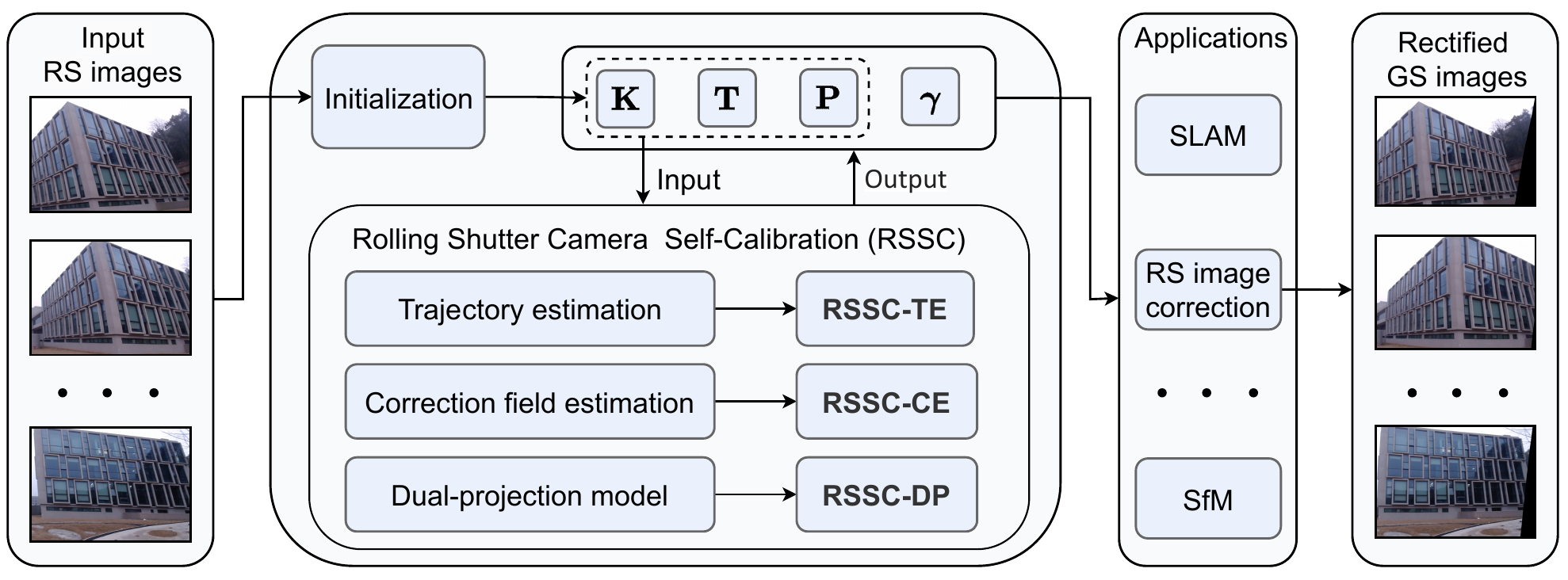}
\caption{Computation pipeline of the proposed Rolling Shutter Camera Self-Calibration (RSSC). Given a sequence of RS images, an initial reconstruction of the camera intrinsics $\mathbf{K}$, camera poses $\mathbf{T}$, and 3D point coordinates $\mathbf{P}$ is first obtained. It is then refined by the RSSC framework, where a self-calibrating BA jointly estimates the readout time ratio $\gamma$. The estimated RS camera parameters can subsequently be used for downstream tasks such as SLAM, SfM, and RS image correction.}
\label{fig: rssc_pipline}
\end{figure}

Although RS camera calibration has been studied, existing methods rely on dedicated calibration patterns or auxiliary sensors, limiting their applicability in unconstrained environments. In practice, the readout time ratio $\gamma$ is often missing, and purely vision-based self-calibration—without calibration targets or additional hardware—remains an open objective. To address this challenge, we propose a general RS self-calibration framework that jointly estimates camera intrinsics and the readout time ratio directly from image observations, as illustrated in Fig.~\ref{fig: rssc_pipline}.

\section{Related Work and Contributions}
\label{sec:Related Works}

\noindent\textbf{Global shutter camera self-calibration.}  Self-calibration for GS cameras has been extensively studied. Methods fall into three groups: classical, deep learning-based, and hybrid methods.

\begin{table}[t]
\centering
\caption{Comparison of representative methods, with applicable sensor type (GS or RS), the need for calibration targets or IMU, and the ability to estimate shutter parameters.}
\resizebox{\textwidth}{!}{  
\begin{tabular}{lcccccc}
\toprule[1pt]
 & \multicolumn{1}{c}{DroidCalib \cite{hagemann2023deep}} 
 & \multicolumn{1}{c}{SelfSup-Calib \cite{fang2022self}} 
 & \multicolumn{1}{c}{ARSC \cite{huai2023automated}} 
 & \multicolumn{1}{c}{RSCC \cite{oth2013rolling}} 
 & \multicolumn{1}{c}{RSC-IMU \cite{huai2022continuous}} 
 & \multicolumn{1}{c}{\textbf{Ours}} \\
\midrule[0.6pt]
Sensor type & GS & GS & RS & RS & RS & RS \\
Target-free & \ding{51} & \ding{51} & \ding{55} & \ding{55} & \ding{51} & \ding{51} \\
IMU-free & \ding{51} & \ding{51} & \ding{51} & \ding{51} & \ding{55} & \ding{51} \\
Shutter param. & \ding{55} & \ding{55} & \ding{51} & \ding{51} & \ding{51} & \ding{51} \\
\bottomrule[1pt]
\end{tabular}
\label{tab:comparision_of_methods}
}
\end{table}

\noindent\textit{Classical self-calibration.} Classical methods are primarily based on multi-view geometric constraints to model image sequences. Early work estimated the intrinsic parameters of a pinhole camera from three views using the Kruppa equations~\cite{maybank1992theory,luong1997self}. Triggs \textit{et al.}~\cite{triggs1998autocalibration} further derived self-calibration constraints for planar scenes. These methods were progressively refined over time, and a comprehensive survey is provided in~\cite{hemayed2003survey}. More recent studies typically adopt motion-based pipelines~\cite{triggs1999bundle,schonberger2016structure}, in which camera intrinsics, 3D structure, and poses are jointly optimized within BA.

\noindent\textit{Learning-based self-calibration.} Deep learning-based methods can be grouped into single-image and video-based methods. Single-image methods include image undistortion networks~\cite{yin2018fisheyerecnet,xue2019learning} and models that directly regress focal length, radial distortion, and horizon parameters~\cite{lee2021ctrl,bogdan2018deepcalib,lopez2019deep,zhuang2019degeneracy}. Video-based methods typically adopt self-supervised learning to jointly regress pixel-wise depth, camera motion, and intrinsic parameters~\cite{gordon2019depth,chen2019self,jeong2021self,fang2022self}. However, the accuracy of the estimated intrinsics is often limited~\cite{gordon2019depth}.

\noindent\textit{Hybrid self-calibration.} Recent works integrate learning with classical BA, which improves robustness and accuracy. BA-Net~\cite{tang2018ba} introduces differentiable BA for end-to-end pose estimation, while DroidCalib~\cite{hagemann2023deep} incorporates a self-calibrating BA layer into learning-based models. VGG-SfM~\cite{wang2024vggsfm} and FlowMap~\cite{smith2025flowmap} further tighten the coupling between neural networks and geometric optimization, jointly enforcing multi-view consistency and optimizing poses, intrinsics, and depth or 3D points within differentiable frameworks.

\noindent\textbf{Rolling shutter camera calibration.}  
Despite its maturity for GS cameras, self-calibration does not exist for RS cameras; instead calibration methods are used, which typically require external calibration targets or additional hardware.

\noindent\textit{RS calibration from calibration board.} Geyer \textit{et al.}~\cite{meingast2005geometric} proposed a geometric model for RS cameras, highlighting their differences from GS cameras and the need for precise LED frequency control for calibration. Huai \textit{et al.}~\cite{huai2023automated} extended this to full intrinsic calibration using an LED panel, estimating both intrinsics and row delay. Oth \textit{et al.}~\cite{oth2013rolling} introduced a method using calibration board videos, jointly estimating intrinsics and row delay via continuous-time trajectory modeling and batch optimization.

\noindent\textit{RS calibration from IMU.} Lee \textit{et al.}~\cite{lee2018calibration} proposed a joint IMU–RS camera calibration using a gray-box framework and an unscented Kalman filter to estimate IMU noise and sensor parameters. Huai \textit{et al.}~\cite{huai2022continuous} introduced a continuous-time B-spline representation, assigning a unique camera pose to each feature, improving calibration accuracy. 

\begin{figure}[t]
\centering
\includegraphics[width=1\columnwidth]{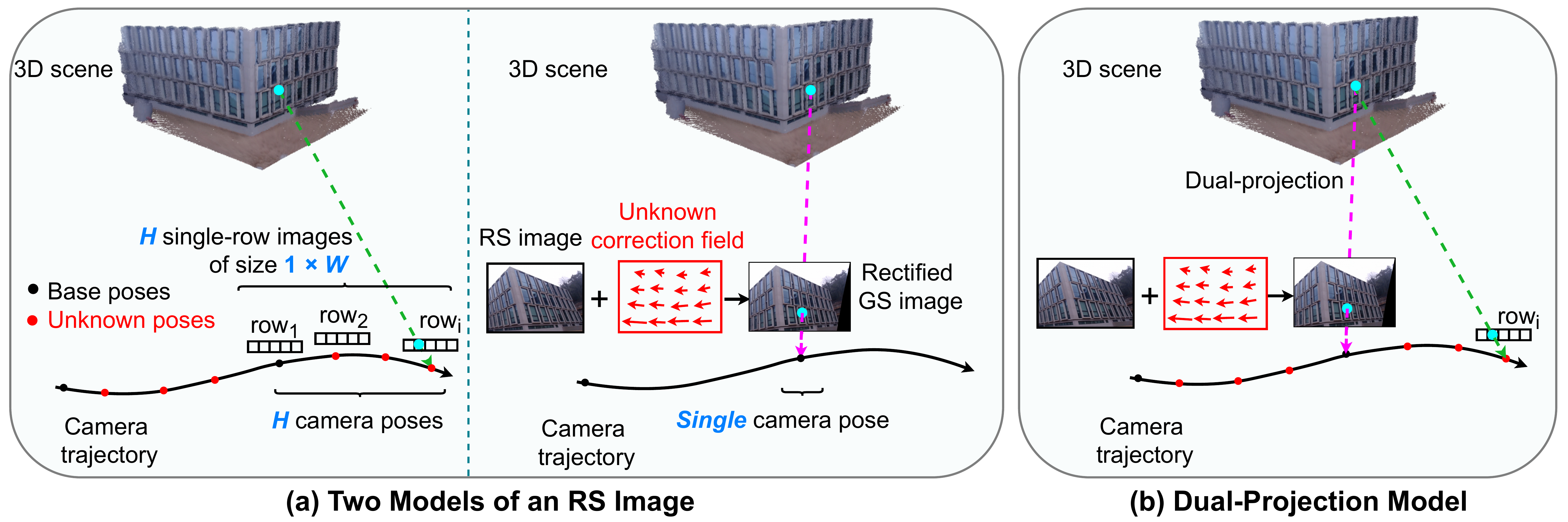}
\caption{(a) Two perspectives for modeling RS images. Left: the RS image is modeled as a composition of multiple independently captured single-row GS images; Right: the RS image is modeled as a single GS image with distortion. (b) The proposed dual-projection model projects each 3D point twice, onto its original RS 2D point and onto the rectified GS 2D point.}
\label{fig: different_RS_view}
\end{figure}

\noindent\textbf{Contributions.}
We propose an RS self-calibration framework where the imaging process is modeled from two complementary perspectives, resulting in three implementations, as illustrated in Fig.~\ref{fig: different_RS_view}.
The first model, denoted RSSC-TE, is based on continuous-time Trajectory Estimation with cumulative B-splines, where camera intrinsics, the readout time ratio, camera poses and 3D points are jointly optimized within a row-wise pose formulation. 
The second model, denoted RSSC-CE, is based on Correction fields Estimation to rectify RS images, followed by a GS BA with an adapted cost. The estimation of correction fields is performed by fitting projected points with a smooth distortion model. 
Building upon these complementary models, we further introduce RSSC-DP, based on a Dual-Projection model in which each 3D point is simultaneously constrained by its original RS observation and its rectified GS counterpart. This unified formulation enforces stronger geometric consistency and improves optimization stability. A comparison with representative methods is provided in Table~\ref{tab:comparision_of_methods}. Our contributions are summarized as follows:
\begin{enumerate}
    \item The first self-calibration method for RS cameras, jointly estimating the intrinsics and the readout time ratio without markers or special hardware.
    
    \item The adaptation and combination of  two RS camera models to enable self-calibration, namely pose-based continuous-time modeling and rectification-based observation modeling. The combination produces a dual-projection formulation, improving stability and consistency.

\end{enumerate}

\section{Background}
\label{sec:Background}
We review the RS projection model and then cumulative B-splines on Lie groups.

\subsection{Rolling Shutter Camera Projection Models}
\label{sec:Rolling Shutter Camera Projection model}
\begin{figure}[t]
\centering
\includegraphics[width=0.95\columnwidth]{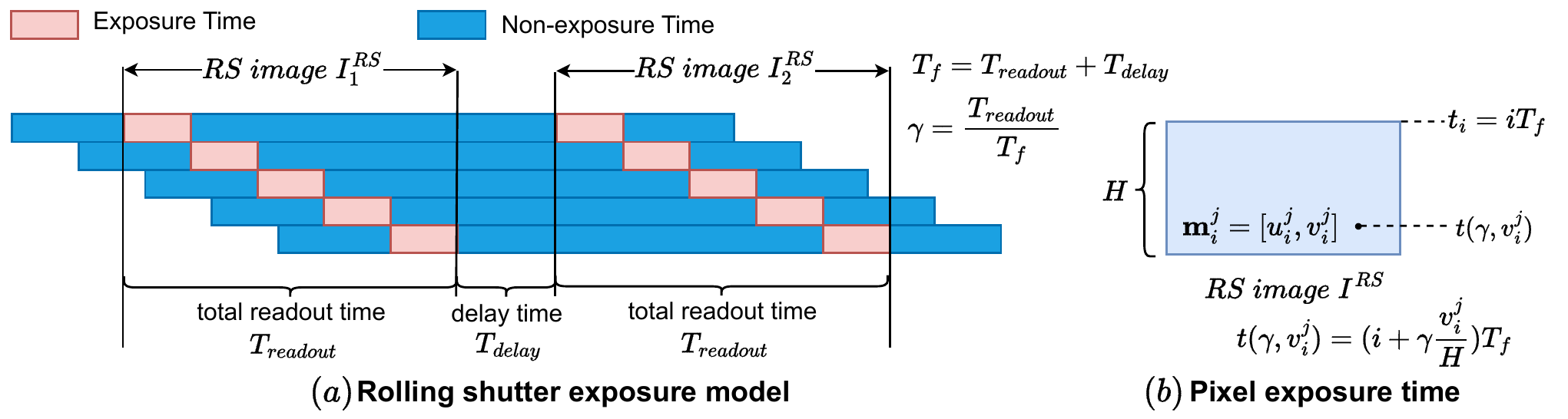}
\caption{Illustration of the RS imaging and pixel-wise exposure modeling. (a) Exposure mechanism between two RS consecutive frames.
(b) Pixel exposure time formulation within an RS image.}
\label{fig: RS exposure}
\end{figure}

Unlike a GS camera, an RS camera captures images row by row. As illustrated in Fig.~\ref{fig: RS exposure}(a), we denote the total readout time and the delay time as $T_{\text{readout}}$ and $T_{\text{delay}}$, respectively, and $T_{f}=T_{\text{readout}}+T_{\text{delay}}$ as the total time interval between two consecutive frames. The readout time ratio $\gamma={T_{\text{readout}}}/{T_f}$ can be used to determine the timestamp of different image rows. As shown in Fig.~\ref{fig: RS exposure}(b), assuming that a 3D point $\mathbf{P}^j = [X^{j}, Y^{j}, Z^{j}]^{\top}$ is observed by an RS camera $i$ represented by measurement $\mathbf{m}_i^j = [u_{i}^{j}, v_{i}^{j}]$ in the image domain, the timestamp of the first row of the $i$-th image is given by $t_{i}=iT_{f}$. 
For an image of height $H$, the timestamp corresponding to the point $\mathbf{m}_{i}^{j}$ can be expressed as:
\begin{equation}\label{equ:rs_timestamp}
        t(\gamma,v_{i}^{j})=\left(i+\gamma \frac{v^{j}_i}{H}\right)T_{f}.
\end{equation}
Different rows in an RS image correspond to different timestamps, so representing the entire image with a single camera pose is inappropriate. Instead, each image row is associated with a distinct camera pose along the continuous motion trajectory. Specifically, for a point $\mathbf{m}_{i}^{j}$ located on the $i$-th RS image, the camera pose is $\mathbf{T}(i, v_i^{j}, \gamma)$. Given intrinsic parameters $\mathbf{K}(f_x, f_y, c_x, c_y)$ and denoting perspective projection as function $\operatorname{\pi}(\cdot)$, RS projection is:
\begin{equation}\label{equ:rs_projection}
        \mathbf{m}_{i}^{j} = \operatorname{\operatorname{\pi}}(\mathbf{K} \mathbf{T}(i, v_i^{j}, \gamma) \mathbf{P}^j).
\end{equation}

\subsection{Cumulative B-Splines on Lie Groups}
\label{sec:Cumulative B-Splines on Lie Groups}
We represent the continuous-time trajectory $\mathbf{T}(t)$ using cumulative cubic B-splines~\cite{lovegrove2013spline}. The trajectory is defined over the time interval $[t_{\text{min}}, t_{\text{max}}]$. 
Its shape is governed by a set of $n$ control poses 
$\{ \mathbf{T}_{i} \in \mathrm{SE}(3) \mid i = 0, \ldots, n-1 \}$, 
which are associated with uniformly spaced knots 
$t_0, t_1, \ldots, t_{n-1}$ within the interval. 
The spacing between consecutive knots is denoted by $\mathrm{\Delta} t$.

Each point along the trajectory is influenced by four consecutive control points. Specifically, for a given time $t \in [t_i, t_{i+1})$, the influencing control points correspond to knots ${t_{i-1}, t_i, t_{i+1}, t_{i+2}}$. The pose at time $t$ is then given by: 
\begin{equation}\label{equ:B-spline-Originally}
\mathbf{T}(t) = \mathbf{T}_{i-1} \prod_{k=1}^{3} \exp\bigl( \mathbf{B}_k(s) \boldsymbol{\Omega}_{i-1+k} \bigr), 
\end{equation}
where $\boldsymbol{\Omega}_i = \log\bigl( \mathbf{T}_{i-1}^{-1} \mathbf{T}_{i} \bigr) \in \mathfrak{se}(3)$ and $s = (t - t_i) / \mathrm{\Delta} t \in [0, 1)$ normalizes the time offset within the current knot interval. The cumulative basis functions $\mathbf{B}(s)$ are:
\begin{equation}\label{equ:B-spline-matrix}
\mathbf{B}(s)
=
\frac{1}{6}
\begin{pmatrix}
6 & 0 & 0 & 0 \\
5 & 3 & -3 & 1 \\
1 & 3 & 3 & -2 \\
0 & 0 & 0 & 1
\end{pmatrix}
\begin{pmatrix}
1 \\
s \\
s^2 \\
s^3
\end{pmatrix},
\end{equation}
where $\mathbf{B}_k(s)$ denotes the $k$-th element of $\mathbf{B}(s)$, with indexing starting from $0$. The resulting pose $\mathbf{T}(t)$ transforms points from local coordinates to world coordinates.

\section{Proposed Method}
\label{sec:Methodology}
We first model an RS image from two complementary perspectives, as illustrated in Fig.~\ref{fig: different_RS_view}(a), from which we derive two RS self-calibration methods. 
We then integrate these models into a unified dual-projection model, as shown in Fig.~\ref{fig: different_RS_view}(b), which jointly enforces motion continuity and observation consistency. 

\begin{figure}[t]
\centering
\includegraphics[width=0.95\columnwidth]{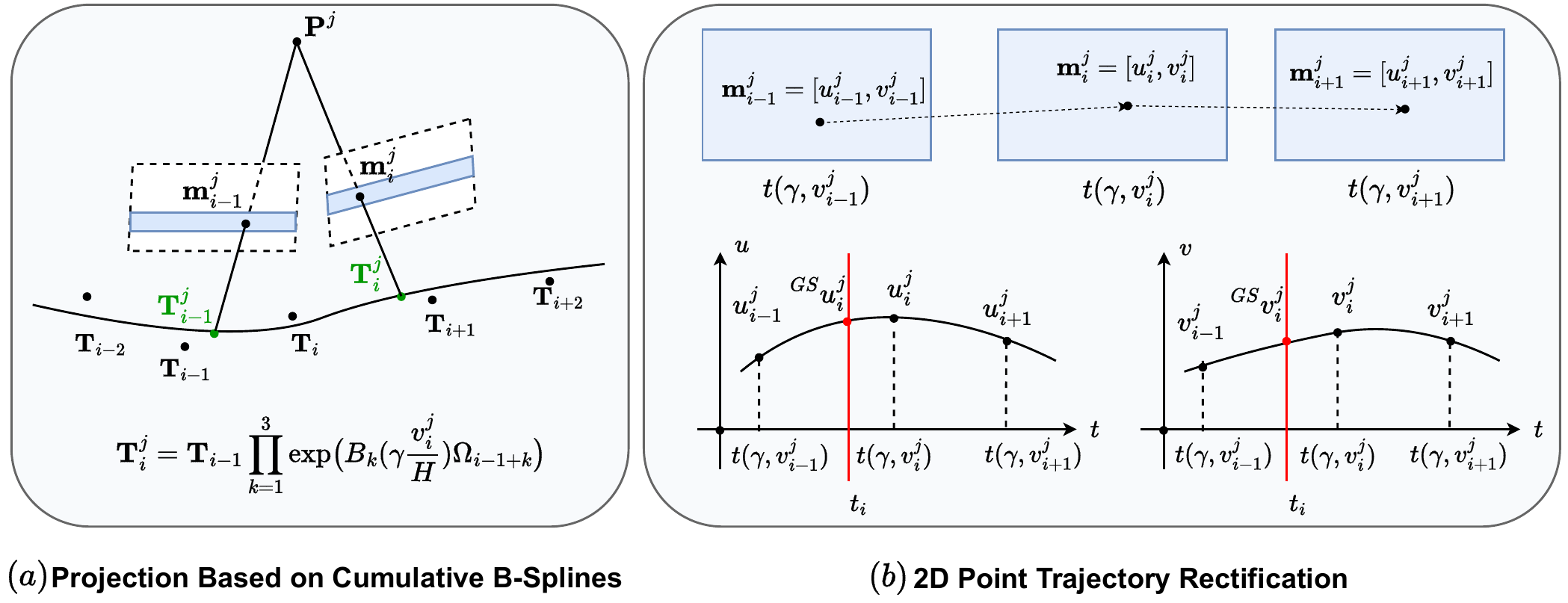}
\caption{Two different fitting strategies:
(a) Representing the camera trajectory as a function of $\gamma$.
(b) Representing the 2D points as a function of $\gamma$.}
\label{fig: projection model}
\end{figure}

\subsection{Continuous-time Trajection Estimation: RSSC-TE}
\label{sec:RSSC-TE}
We model an RS image as a collection of $H$ individual single-row GS images (Fig.~\ref{fig: different_RS_view}(a), left) and introduce RSSC-TE, RS self-calibration via continuous-time trajectory estimation with cumulative B-splines.

\noindent\textbf{Continuous-time pose parameterization.}
\label{sec:Continuous-Time Pose Parameterization}
As illustrated in Fig.~\ref{fig: projection model}(a), each row of the image corresponds to a different camera pose, and the camera trajectory is modeled using cumulative B-splines as described in Sec.~\ref{sec:Background}.
Assuming that $n$ images are captured, we introduce $n$ control poses $\{ \mathbf{T}_i \in \mathrm{SE}(3) \mid i = 0, \ldots, n-1 \}$, which parameterize the continuous camera trajectory over time. The temporal interval between two consecutive control poses is uniformly equal to $T_f$.

For the $i$-th image, we denote the pose corresponding to row $v_i^j$ as $\mathbf{T}_i^j$. This pose is influenced by four consecutive control poses, namely $\mathbf{T}_{i-1}$, $\mathbf{T}_{i}$, $\mathbf{T}_{i+1}$, and $\mathbf{T}_{i+2}$. Their associated timestamps are $(i-1)T_f$, $iT_f$, $(i+1)T_f$, and $(i+2)T_f$, respectively. The timestamp of row $v_i^j$ depends on $\gamma$ and is given by $t(\gamma,v_{i}^{j}) = (i+\gamma \frac{v_i^j}{H})T_f$.
According to Eq.~(\ref{equ:B-spline-Originally}), under the assumption of a reasonably smooth camera trajectory, the pose corresponding to row $v_i$ is: 
\begin{equation}\label{equ:B-spline}
\mathbf{T}_i^j = \mathbf{T}_{i-1} \prod_{k=1}^{3} \exp\bigl( \mathbf{B}_k(\gamma \frac{v_i^j}{H}) \boldsymbol{\Omega}_{i-1+k} \bigr).
\end{equation}
\noindent\textbf{Error function.}
\label{Error Function of RSSC-TE}
The non-linear least squares optimizers are used to find a solution $\boldsymbol{\theta}^*_{\text{TE}}$ including readout time ratio $\gamma^*$, camera intrinsic parameters $\mathbf{K}^*$, camera poses $\mathbf{T}^*$ and 3D points $\mathbf{P}^*$ by minimizing the reprojection error  from point $j$ to camera $i$ over all the camera index in set $\mathcal{F}$ and corresponding 3D points index in subset $\mathcal{P}_i$: 
\begin{equation}
    \begin{aligned}
        \boldsymbol{\theta}^*_{\text{TE}} =  \left \{\gamma^*, \mathbf{K}^*, \mathbf{T}^*, \mathbf{P}^* \right \} = \mathop{\arg\min}_{\boldsymbol{\theta}_{\text{TE}} }  \sum_{i\in \mathcal{F}}\sum_{j\in \mathcal{P}_i }  \left \| {}^{TE}\mathbf{e}_{i}^{j} \right \|^{2}_{2}
    \end{aligned},
    \label{equ:cost_function}
\end{equation}
where the reprojection residual ${}^{\text{TE}}\mathbf{e}^j_i$ is expressed as:
\begin{equation}
    \begin{aligned}
{}^{\text{TE}}\mathbf{e}_i^j = \mathbf{m}_i^j -\operatorname{\pi}\left(\mathbf{K} \mathbf{T}_{i-1} \prod_{k=1}^{3} \exp\left( \mathbf{B}_k(\gamma \frac{v_i^j}{H}) \boldsymbol{\Omega}_{i-1+k} \right) \mathbf{P}^j\right)
    \end{aligned}.
    \label{equ:error_function}
\end{equation}

\subsection{Correction Field Estimation: RSSC-CE}
\label{sec:RSSC-CE}
We model an RS image as a distorted GS image associated with an unknown correction field (Fig.~\ref{fig: different_RS_view}(a), right) and develop RSSC-CE, an RS self-calibration method based on correction field estimation.

\noindent\textbf{Correction field estimation using 2D point trajectory fitting.}
\label{sec:2D Pixel Points Trajectory Fitting} 
As illustrated in Fig.~\ref{fig: projection model}(b), we model the trajectory of each 2D point as a function of time and temporally rectify observations to align them to a unified GS timestamp.
For a 3D point $\mathbf{P}^j$, its 2D observation in frame $i$ is denoted as $\mathbf{m}_{i}^{j}$ with timestamp $t(\gamma, v_{i}^{j})$ (Eq.~(\ref{equ:rs_timestamp})). Given a target GS time $t_i$, we estimate the rectified point location ${}^{\text{GS}}\mathbf{m}_{i}^{j}$ by fitting a local trajectory and evaluating it at $t_i$. We consider two interpolation schemes for this purpose.
Both schemes can be used to approximate and rectify RS pixel trajectories, and their performance will be compared in the experiments.

\noindent\textit{Quadratic Polynomial Interpolation.}
Following~\cite{qu2023towards}, a quadratic polynomial is fitted using three temporally adjacent observations from frames $i\!-\!1$, $i$, and $i\!+\!1$.
The resulting method is referred to as RSSC-CEQ.
The rectified point is obtained via Lagrange interpolation:
\begin{equation}
{}^{\text{GS}}\mathbf{m}_{i}^{j}(t_i) =
L_0(t_i) \mathbf{m}_{i-1}^{j} +
L_1(t_i) \mathbf{m}_{i}^{j} +
L_2(t_i) \mathbf{m}_{i+1}^{j},
\end{equation}
where the Lagrange basis functions are:
\begin{equation}
\begin{cases}
L_0(t_{i}) = \frac{(t_i - t(\gamma, v_{i}^{j}))(t_i - t(\gamma, v_{i+1}^{j}))}{(t(\gamma, v_{i-1}^{j}) - t(\gamma, v_{i}^{j}))(t(\gamma, v_{i-1}^{j}) - t(\gamma, v_{i+1}^{j}))},\\
L_1(t_{i}) = \frac{(t_i - t(\gamma, v_{i-1}^{j}))(t_i - t(\gamma, v_{i+1}^{j}))}{(t(\gamma, v_{i}^{j}) - t(\gamma, v_{i-1}^{j}))(t(\gamma, v_{i}^{j}) - t(\gamma, v_{i+1}^{j}))},\\
L_2(t_{i}) = \frac{(t_i - t(\gamma, v_{i-1}^{j}))(t_i - t(\gamma, v_{i}^{j}))}{(t(\gamma, v_{i+1}^{j}) - t(\gamma, v_{i-1}^{j}))(t(\gamma, v_{i+1}^{j}) - t(\gamma, v_{i}^{j}))}.
\end{cases}
\end{equation}

\noindent\textit{Cubic Hermite Interpolation.}
We also explore a cubic Hermite interpolation scheme using four consecutive observations from frames $i\!-\!1$, $i$, $i\!+\!1$, and $i\!+\!2$. 
The resulting method is referred to as RSSC-CEH.
The rectified point is:
\begin{equation}
{}^{\text{GS}}\mathbf{m}_{i}^{j}(t_i) =
h_{00}(\tau)\mathbf{m}_{i}^{j}
+ h_{10}(\tau)\mathrm{\Delta} t\, \mathbf{d}_{i}
+ h_{01}(\tau)\mathbf{m}_{i+1}^{j}
+ h_{11}(\tau)\mathrm{\Delta} t\, \mathbf{d}_{i+1},
\end{equation}
where $\tau$ is the interpolation parameter:
\begin{equation}
\tau = \frac{t_i - t(\gamma, v_{i}^{j})}{t(\gamma, v_{i+1}^{j}) - t(\gamma, v_{i}^{j})}, 
\end{equation}
and the Hermite basis functions are:
\begin{equation}
\begin{cases}
\begin{aligned}
h_{00}(\tau) &= 2\tau^3 - 3\tau^2 + 1,\\
h_{10}(\tau) &= \tau^3 - 2\tau^2 + \tau,\\
h_{01}(\tau) &= -2\tau^3 + 3\tau^2,\\
h_{11}(\tau) &= \tau^3 - \tau^2.
\end{aligned}
\end{cases}
\end{equation}
The tangent vectors are approximated using central differences:
\begin{equation}
\mathbf{d}_{i} = 
\frac{\mathbf{m}_{i+1}^{j} - \mathbf{m}_{i-1}^{j}}
{t(\gamma, v_{i+1}^{j}) - t(\gamma, v_{i-1}^{j})}, \quad
\mathbf{d}_{i+1} = 
\frac{\mathbf{m}_{i+2}^{j} - \mathbf{m}_{i}^{j}}
{t(\gamma, v_{i+2}^{j}) - t(\gamma, v_{i}^{j})}.
\end{equation}

\noindent\textbf{Error function.}
\label{Error Function of RSSC-CE}
Following Sec.~\ref{sec:RSSC-TE}, we estimate the parameters $\boldsymbol{\theta}^*_{\text{CE}}$
via non-linear least squares optimization.
The cost is the reprojection error over all camera indices in $\mathcal{F}$ and corresponding 3D point indices in $\mathcal{P}_i$:
\begin{equation}
    \begin{aligned}
        \boldsymbol{\theta}^*_{\text{CE}} =  \left \{\gamma^*, \mathbf{K}^*, {}^{\text{GS}}\mathbf{T}^*, \mathbf{P}^* \right \} = \mathop{\arg\min}_{\boldsymbol{\theta}_{\text{CE}} }  \sum_{i\in \mathcal{F}}\sum_{j\in \mathcal{P}_i }  \left \| {}^{\text{CE}}\mathbf{e}_{i}^{j} \right \|^{2}_{2}
    \end{aligned},
    \label{equ:cost_function}
\end{equation}
where the reprojection residual ${}^{\text{CE}}\mathbf{e}^j_i$ is expressed as:
\begin{equation}
    \begin{aligned}
{}^{\text{CE}}\mathbf{e}_i^j = {}^{\text{GS}}\mathbf{m}_{i}^{j}(t_i) -\operatorname{\pi}(\mathbf{K} {}^{\text{GS}}\mathbf{T}_i \mathbf{P}^j)
    \end{aligned}.
    \label{equ:error_function}
\end{equation}

\subsection{Combined Method: RSSC-DP}
\label{sec:RSSC-DP}
We combine RSSC-TE, which captures motion continuity via continuous-time pose modeling, and RSSC-CE, which enforces observation consistency through 2D points rectification, into a unified dual-projection formulation RSSC-DP (Fig.~\ref{fig: different_RS_view}(b)). Under this model, each 3D point is constrained by both its original RS observation and the corresponding rectified GS observation. As shown in Sec.~\ref{sec:Experiment}, this additional constraint improves the robustness and accuracy of self-calibration.

\noindent\textbf{Continuous-time constrained GS poses.}
In RSSC-CE, the rectified GS pose ${}^{\text{GS}}\mathbf{T}_i$ at timestamp $t_i$ was previously treated as an independent optimization variable. 
In RSSC-DP, we instead parameterize it using the same cumulative B-spline trajectory defined in Sec.~\ref{sec:Continuous-Time Pose Parameterization}. 
Specifically, let $t_i$ denote the reference GS timestamp of frame $i$. 
The corresponding pose is obtained by evaluating the continuous trajectory:
\begin{equation}
\mathbf{T}(t_i) = \mathbf{T}_{i-1} \prod_{k=1}^{3} \exp\bigl( \mathbf{B}_k(s_i) \boldsymbol{\Omega}_{i-1+k} \bigr),
\label{equ:fusion_pose}
\end{equation}
where: 
\[
s_i = \frac{t_i - iT_f}{T_f},
\]
and $\mathbf{T}_{i-1}, \ldots, \mathbf{T}_{i+2}$ are the four control poses influencing $t_i$.
Accordingly, we replace ${}^{\text{GS}}\mathbf{T}_i$ with the B-spline controlled trajectory $\mathbf{T}(t_i)$.
As a result, the rectified GS poses are no longer independent parameters, but are fully determined by the continuous-time trajectory.

\noindent\textbf{Error function.}
With the continuous-time constrained GS poses introduced above, RSSC-TE and RSSC-CE are no longer two independent formulations, but become two complementary projection instances defined over the same underlying trajectory.
For each 3D point $\mathbf{P}^j$ observed in frame $i$, the unified trajectory naturally induces two geometrically related projection events:
1) a row-dependent RS projection at timestamp $t(\gamma, v_i^j)$, and 2) a reference GS projection at timestamp $t_i$.
The RSSC-TE residuals captures consistency with the original RS observation:
\begin{equation}
{}^{\text{TE}}\mathbf{e}_i^j 
=
\mathbf{m}_i^j 
-
\pi\!\left(
\mathbf{K} \mathbf{T}(t(\gamma, v_i^j)) \mathbf{P}^j
\right),
\label{equ:fusion_bot_error}
\end{equation}
while the RSSC-CE residuals captures consistency with its rectified GS counterpart:
\begin{equation}
{}^{\text{CE}}\mathbf{e}_i^j 
=
{}^{\text{GS}}\mathbf{m}_i^j(t_i)
-
\pi\!\left(
\mathbf{K} \mathbf{T}(t_i) \mathbf{P}^j
\right).
\label{equ:fusion_bop_error}
\end{equation}
These two residuals arise from two temporally distinct projections of the same 3D point along a shared continuous trajectory. 
The resulting objective is formulated as a weighted least-squares problem:
\begin{equation}
\boldsymbol{\theta}^*_{\text{DP}}
=
\{
\gamma^*,
\mathbf{K}^*,
\mathbf{T}^*,
\mathbf{P}^*
\}
=
\mathop{\arg\min}_{\boldsymbol{\theta}_{\text{DP}}}
\sum_{i\in \mathcal{F}}
\sum_{j\in \mathcal{P}_i}
\Big(
\| {}^{\text{TE}}\mathbf{e}_i^j \|_2^2
+
\lambda
\| {}^{\text{CE}}\mathbf{e}_i^j \|_2^2
\Big),
\label{equ:fusion_cost}
\end{equation}
where $\lambda$ balances the contributions of the two projection models.

\begin{figure}[t]
\centering
\includegraphics[width=0.9\columnwidth]{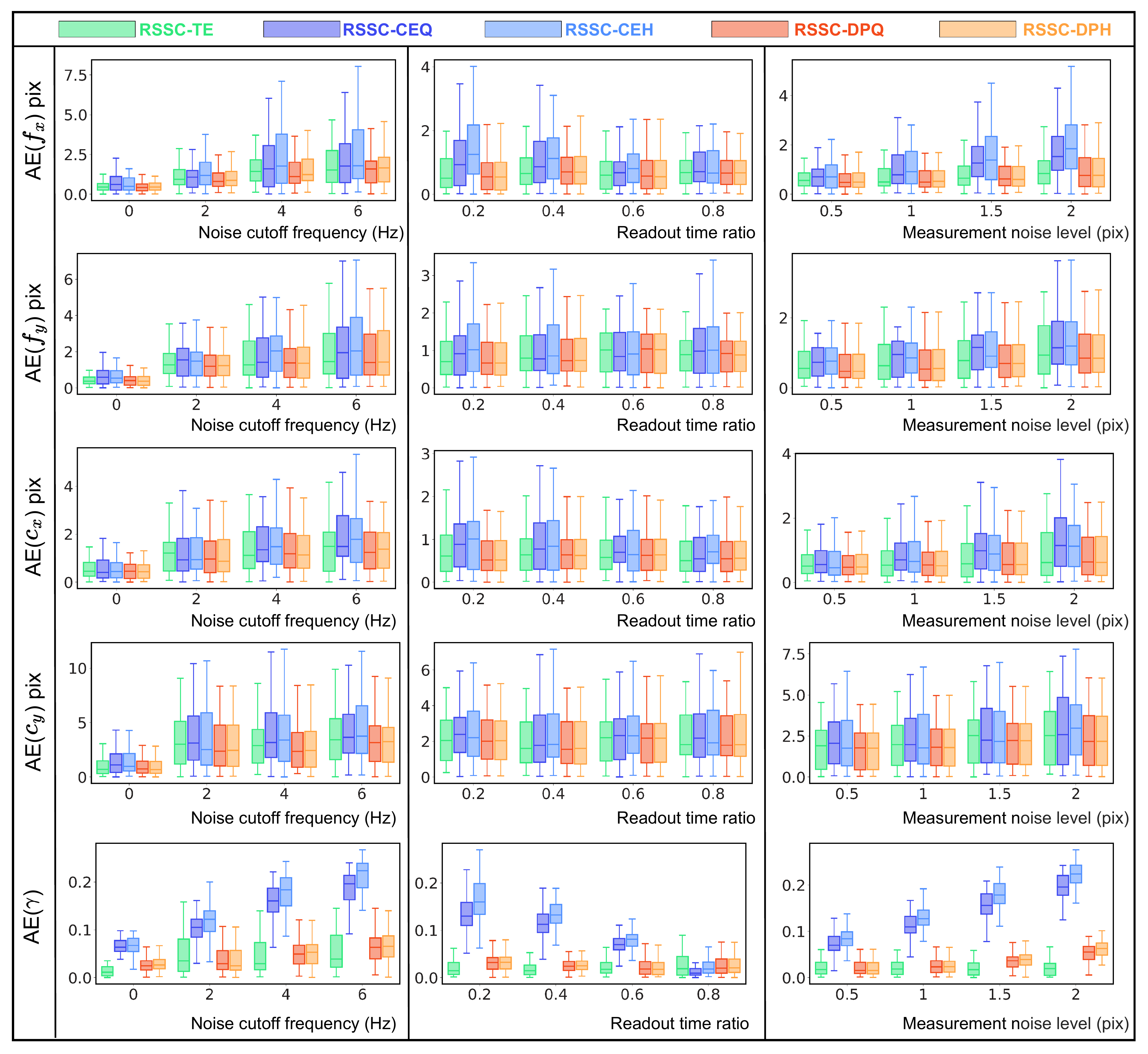}
\caption{Intrinsic parameters and readout time ratio errors for the proposed RS camera self-calibration methods under varying experimental conditions, with, from left to right, trajectory smoothness, readout time ratio, and measurement noise.}
\label{fig: simulate_result}
\end{figure}

\subsection{Discussion}
The two constraints in RSSC-DP are complementary and compatible in practice. A non-smooth 3D trajectory can still yield locally smooth 2D projections, allowing RSSC-CE to compensate for inaccuracies in RSSC-TE's global trajectory representation; conversely, when the 2D motion field is locally irregular, RSSC-TE's global smoothness prior can mitigate per-point fitting errors in RSSC-CE. If the objectives were conflicting, RSSC-DP would be expected to perform substantially worse than at least one individual model in some scenarios. However, as shown in Sec.~\ref{sec:Experiment}, such degradation was never observed, and the performance gap remained small even when RSSC-DP was not the best performer.

\section{Experimental Results}
\label{sec:Experiment}
We first investigate the influence of different conditions on the proposed RSSC methods through simulated data. We then evaluate on real datasets and compare with state-of-the-art methods.

\subsection{Synthetic Data Experiments}
\label{sec:Simulated Data Experiments}

\noindent\textbf{Simulation setup and compared methods.}
The 3D scene is defined within a $4\times4\times4$ cubic volume. For each trial, a smooth continuous-time camera trajectory is generated, and we conduct a total of 50 independent trials, each consisting of 50 images. The weighting parameter $\lambda$ in Eq.~(\ref{equ:fusion_cost}) is set to 0.5.
The proposed methods—RSSC-TE, RSSC-CEQ, RSSC-CEH, RSSC-DPQ, and RSSC-DPH—are evaluated under varying simulation conditions.

\noindent\textbf{Experimental variables.}
We evaluate the proposed methods under three simulation factors, which include trajectory smoothness, RS strength, and measurement noise.

Trajectory smoothness is controlled by injecting band-limited Gaussian noise into the continuous-time trajectory $\mathbf{x}(t) \in \mathbb{R}^6$. The perturbed trajectory is:
\begin{equation}
\tilde{\mathbf{x}}(t) = \mathbf{x}(t) + \tilde{\mathbf{n}}(t), \quad
\tilde{\mathbf{n}}(t) = [\tilde{n}_1(t),\dots,\tilde{n}_6(t)]^\top.
\end{equation}
Let $n_d(t) \sim \mathcal{N}(0,1)$ denote zero-mean unit-variance Gaussian noise. The band-limited Gaussian noise $\tilde{n}_d(t)$ is computed as
\begin{equation}
\tilde{n}_d(t) = \lambda_d \, \mathcal{F}^{-1}\!\left( H(f; f_c, m)\,\mathcal{F}(n_d(t)) \right),
\end{equation}
where $H(f; f_c,m)$ denotes the frequency response of a low-pass Butterworth filter with cutoff frequency $f_c$ and order $m$. The filter order is fixed to $m=4$, and the parameter $\lambda_d$ controls the noise amplitude, set to $0.015$\,rad for rotation and $1.5$ for translation. The cutoff frequency $f_c \in \{0,2,4,6\}\,\text{Hz}$ controls the trajectory smoothness, where larger $f_c$ results in less smooth motion.

RS distortion strength is controlled by the readout time ratio $\gamma \in \{0.2,0.4,0.6,0.8\}$, and measurement noise by the standard deviation $\sigma \in \{0.5,1.0,1.5,2.0\}$ pixels of Gaussian noise added to the 2D measurements.

\begin{table*}[t]
\centering
\caption{Optimization efficiency comparison. The rows correspond to per-iteration time (seconds), iterations to convergence, and total optimization time, respectively. All results are medians over 50 runs. Bold is best and underlined second-best.}
\scalebox{0.95}{
\begin{tabular*}{\textwidth}{@{\extracolsep{\fill}}lccccc}
\toprule[1pt]
Method & RSSC-TE & RSSC-CEQ & RSSC-CEH & RSSC-DPQ & RSSC-DPH \\
\midrule[0.6pt]
Iteration time (s)  & 0.5142 & \textbf{0.0841} & \underline{0.0851} & 0.7668 & 0.7735 \\
Nb. iterations      & 32      & \textbf{5}      & \underline{6}      & 9      & 14      \\
Total time (s)   & 5.8657      & \textbf{0.2321}   & \underline{0.2765}      & 2.6048      & 3.9185      \\
\bottomrule[1pt]
\end{tabular*}
}
\label{tab:computation_time}
\end{table*}

\noindent\textbf{Evaluation metrics.} 
We evaluate the proposed methods using two key quantitative metrics. First, the absolute error (AE) is computed for the camera intrinsic parameters and the readout time ratio, $\theta \in \{f_x, f_y, c_x, c_y, \gamma\}$, defined as $\text{AE}(\hat{\theta}) = |\hat{\theta} - \theta|$, where $\theta$ and $\hat{\theta}$ denote the true and estimated values, respectively. Second, we assess optimization efficiency by measuring the per-iteration time, the number of iterations to convergence, and the total optimization time. Convergence is declared when the parameter update is below $10^{-8}$, the infinity norm of the gradient is below $10^{-10}$, or the change in the objective function is below $10^{-8}$.

\noindent\textbf{Analysis of experimental results.} 
The results of the proposed methods under varying simulation conditions are given in Fig.~\ref{fig: simulate_result} and Table~\ref{tab:computation_time}. 

Regarding trajectory smoothness, estimation errors increase for all methods as trajectory perturbation grows, indicating reduced stability under less smooth motion. The RSSC-DP methods consistently achieve the lowest errors, showing superior robustness from the dual-projection formulation that jointly constrains pose and projection, while RSSC-CE degrades more rapidly due to its sensitivity to motion inaccuracies.

For variations in the readout time ratio $\gamma$, intrinsic parameter estimation remains largely unaffected for all methods. However, RSSC-CE exhibits increasing errors in the estimated $\gamma$ as the readout time decreases, whereas RSSC-TE and RSSC-DP maintain stable performance.

When considering measurement noise, RSSC-CE is highly sensitive, with errors increasing rapidly as 2D noise grows. In contrast, RSSC-TE and RSSC-DP remain robust, benefiting from continuous-time modeling and dual-projection constraints that provide inherent regularization.

In terms of computational efficiency, Table~\ref{tab:computation_time} shows that RSSC-CE achieves the fastest per-iteration time and requires the fewest iterations, yielding the lowest total optimization time. RSSC-TE has slightly shorter per-iteration time than RSSC-DP but requires substantially more iterations, resulting in the highest total time. Overall, RSSC-DP balances accuracy and efficiency, offering competitive performance at moderate computational cost.

Overall, the results highlight that RSSC-TE excels under smooth trajectories, RSSC-CE prioritizes computational efficiency, and RSSC-DP achieves the best accuracy and robustness across all conditions, making it the most versatile choice for practical applications.

\begin{table*}[t]
\centering
\caption{Self-calibration results on WHU-RSVI~\cite{cao2020whu} and TUM-RSVI~\cite{dai2016rolling} datasets. All results are median absolute errors (MAE) of the intrinsic parameters and readout time ratio, and median absolute trajectory errors (MATE) over all sequences in each dataset.}
\scalebox{0.95}[0.95]{
\begin{tabular}{l l c c c c c c}
\Xhline{1pt}
Dataset & Method & MAE($f_x$)& MAE($f_y$) & MAE($c_x$) & MAE($c_y$)  & MAE($\gamma$) & MATE \\
\Xhline{1pt}

\multirow{8}{*}{WHU}
& COLMAP~\cite{schonberger2016structure}   & 3.05 & 11.26  & 4.36 & 8.17 & --- & 0.0453 \\
& DroidCalib~\cite{hagemann2023deep}    & 0.56  & 22.33  & 3.13  & 13.80  & ---  & 0.0317  \\
& SelfSup-Calib~\cite{fang2022self} & 195.98  & 129.28  & 13.85  & 23.78  & ---  & 0.4585  \\
& RSSC-TE     & \underline{0.07}  & \textbf{0.31}  & \underline{0.23}  & \textbf{0.60}  & \underline{0.0485}  & 0.0170  \\
& RSSC-CEQ    & 0.16  & 2.82  & 0.27  & 1.04  & 0.0814  & 0.0131  \\
& RSSC-CEH    & 0.67  & 2.43  & 0.71  & 1.60  & 0.1789  & 0.0238  \\
& RSSC-DPQ   & \textbf{0.03}  & \underline{0.44}  & \textbf{0.19}  & \underline{0.65}  & \textbf{0.0483}  & \textbf{0.0101}  \\
& RSSC-DPH   & 0.30  & \textbf{0.31}  & 0.24  & 1.50  & 0.0945  & \underline{0.0126}  \\

\midrule

\multirow{8}{*}{TUM}
& COLMAP~\cite{schonberger2016structure}   & 33.58 &	14.82 & 16.01 & 28.5 & -- &	0.0374
  \\
& DroidCalib~\cite{hagemann2023deep}    & 71.96 & 67.56 & 17.54 & 41.76 & -- & 0.0963
  \\
& SelfSup-Calib~\cite{fang2022self} & 359.77 & 338.17 & 59.885 & 9.85 & -- & 0.1432
  \\
& RSSC-TE      & \underline{3.16} & \textbf{1.48} & 1.76 & \textbf{1.12} & 0.1144 & \underline{0.0063}
  \\
& RSSC-CEQ    & 3.79 & 2.67 & \underline{1.71} & 1.25 & \textbf{0.0609} & 0.0066
  \\
& RSSC-CEH    & 3.71 & 3.31 & \textbf{1.40} & 1.52 & 0.0852 & 0.0081
  \\
& RSSC-DPQ   & \textbf{3.10} & \underline{2.39} & 1.93 & \underline{1.16} & \underline{0.0814} & \textbf{0.0055}  \\
& RSSC-DPH   & 4.17	& 3.06 & 2.16 & 2.14 & 0.1374 & 0.0072
 \\

\Xhline{1pt}
\end{tabular}
}
\label{table:self-calibration-results}
\end{table*}

\subsection{Real Data Experiments}
\label{sec:Real Data Experiments}

\begin{figure}[http]
\centering
\includegraphics[width=1\columnwidth]{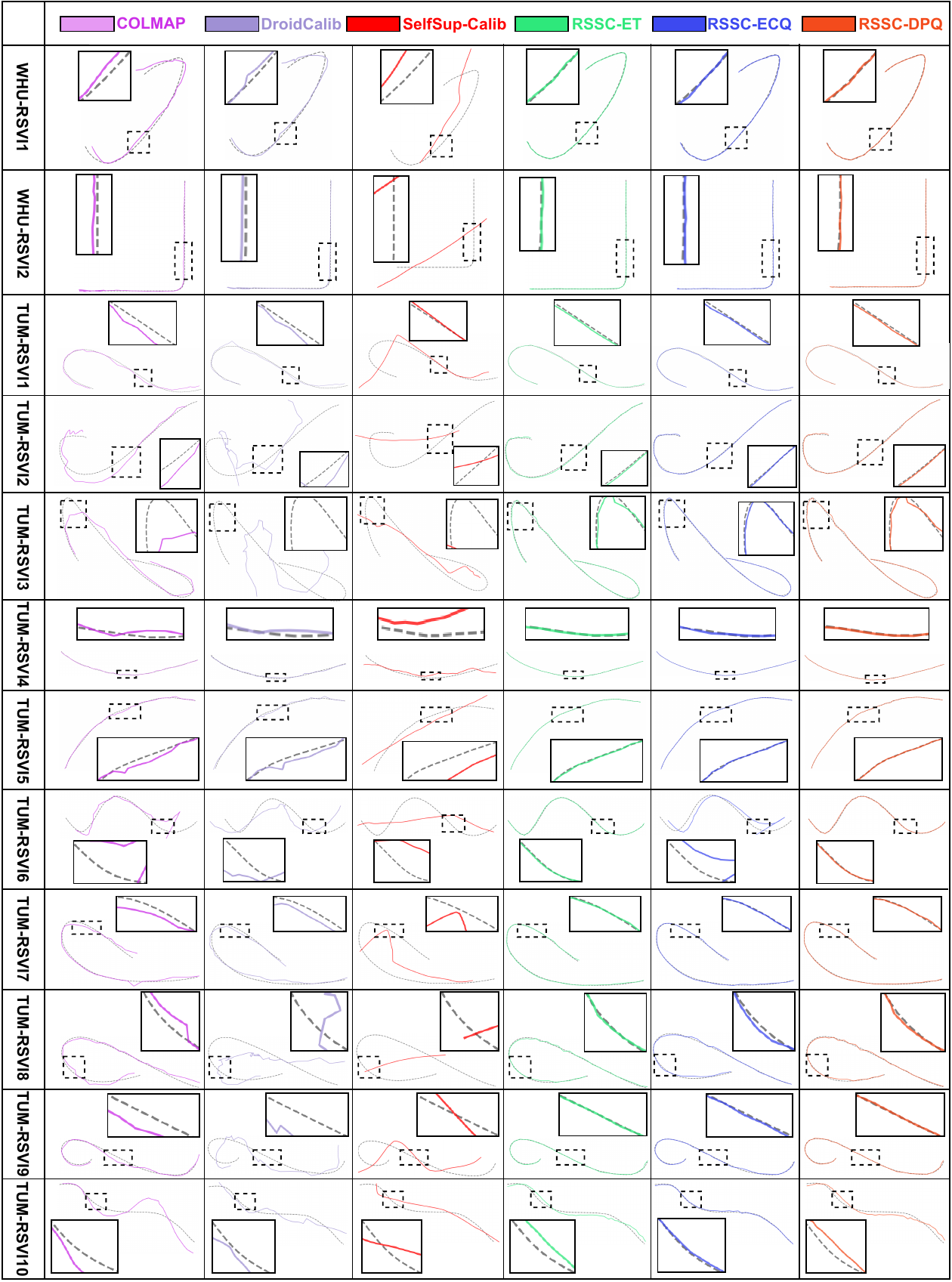}
\caption{Trajectory comparison on the WHU-RSVI~\cite{cao2020whu} and TUM-RSVI~\cite{dai2016rolling} datasets. Each row denotes a different trajectory sequence, while each column corresponds to a specific method. Selected local regions (highlighted by dashed boxes) are magnified and displayed within solid boxes for detailed inspection.}
\label{fig: real_data_result}
\end{figure}

\noindent\textbf{Datasets and compared methods.} 
We conduct self-calibration evaluation on two publicly available RS datasets, WHU-RSVI~\cite{cao2020whu} and TUM-RSVI~\cite{dai2016rolling}. For each sequence, 50 consecutive frames are selected for calibration. Our method uses COLMAP~\cite{schonberger2016structure} to obtain an initial reconstruction, with the readout time ratio initialized to 0.5, and the weighting parameter $\lambda$ in Eq.~(\ref{equ:fusion_cost}) is set to 0.5. The comparison methods include \textit{1)} COLMAP~\cite{schonberger2016structure}, \textit{2)} DroidCalib~\cite{hagemann2023deep}, and \textit{3)} SelfSup-Calib~\cite{fang2022self}. Since no publicly available methods exist for self-calibration of RS cameras, only approaches designed for GS cameras are included in the comparison.

\noindent\textbf{Evaluation metrics.}
We evaluate the accuracy of intrinsic parameters and the readout time ratio using the median absolute error (MAE), together with the median absolute trajectory error (MATE) after alignment to the ground-truth trajectory using a $\mathrm{Sim}(3)$ similarity transformation.

\noindent\textbf{Experiment results.} 
Table~\ref{table:self-calibration-results} presents the intrinsic parameters and readout time ratio MAEs, and MATE computed over all sequences in each dataset, and Fig.~\ref{fig: real_data_result} shows a comparison between the trajectories obtained by different methods and the ground truth (full results are provided in the supplementary material). We observe that, for all evaluation metrics, our self-calibration methods consistently outperform the baselines. Among the proposed methods, RSSC-TE and RSSC-DPQ achieve the best performance. These results clearly validate the effectiveness, accuracy, and robustness of the proposed RS self-calibration approach. Additional comparisons with target-based RS calibration methods are provided in the supplementary material, where our approach achieves comparable performance.

\section{Conclusion}
We have proposed the first RS camera self-calibration framework, jointly estimating intrinsics and the readout time ratio from a target-free general video. Our approach models self-calibration from two complementary perspectives: RSSC-TE performs trajectory estimation using continuous-time B-splines to optimize intrinsics, readout time ratio, poses and 3D points; RSSC-CE estimates correction fields by fitting 2D pixel trajectories to rectify RS images into a GS formulation. These strategies are unified in a dual-projection model, RSSC-DP, which enforces geometric consistency between row-dependent and rectified observations, thereby enhancing both estimation accuracy and robustness under challenging motion and imaging conditions. Extensive experiments on synthetic and real data demonstrate the accuracy, robustness, and practical effectiveness of our method. Future work includes jointly optimizing RS camera parameters and dense 3D scene representations, such as 3D Gaussian models, as well as studying degeneracies in RS camera self-calibration.

\par\vfill\par



%
%
\bibliographystyle{splncs04}
\bibliography{main}
\end{document}


\title{Supplementary Material \\ Rolling Shutter Camera Self-Calibration} 

\titlerunning{Supplementary Material \\ Rolling Shutter Camera Self-Calibration}

\author{Yongcong Zhang\inst{1,2,*}\orcidlink{0009-0009-1254-6549} \and 
Navid Rabbani\inst{1,*}\orcidlink{0009-0006-2409-4824} \and
Bangyan Liao\inst{3,*}\orcidlink{0009-0007-7739-4879} \and \\
Chengbo Wang\inst{4}\orcidlink{0009-0009-1385-1633} \and 
Yizhen Lao\inst{4}\orcidlink{0000-0002-6284-1724} \and
Adrien Bartoli\inst{1,2,\dagger}\orcidlink{0000-0003-3545-7329}
}   

\authorrunning{Y. Zhang, N. Rabbani et al.}


\institute{
\textsuperscript{1}{DIA2M, DRCI, CHU Clermont-Ferrand, France} \\
\textsuperscript{2}SURGAR, Surgical Augmented Reality, Clermont-Ferrand, France \\
\textsuperscript{3}School of Engineering, Westlake University, China \\ 
\textsuperscript{4}School of Design, Hunan University, China
}
\maketitle
\renewcommand{\thefootnote}{\relax}\footnotetext{
\textsuperscript{$\star$} Equal contribution\\  
\inst{\dagger} Corresponding author: \email{adrien.bartoli@gmail.com} \\
Project page: \href{https://github.com/Yongcong-Zhang/RSSC}{https://github.com/Yongcong-Zhang/RSSC}}
\renewcommand{\thefootnote}{\arabic{footnote}}
\setcounter{footnote}{0}


\section*{Overview}
In this supplementary material, we further discuss the following content.
\begin{itemize}
    \item Algorithmic Summary (Sec.~\ref{supp:sec:Algorithmic Summary}).
    \item Jacobian Derivation (Sec.~\ref{supp:sec:Jacobian Derivation for RSSC}).
    \begin{itemize}
        \item Jacobian Derivation for RSSC-TE (Sec.~\ref{supp:sec:Jacobian Derivation for RSSC-TE}).
        \item Jacobian Derivation for RSSC-CE (Sec.~\ref{supp:sec:Jacobian Derivation for RSSC-CE}).
    \end{itemize}
    \item Complete Real Data Experimental Results (Sec.~\ref{supp:sec:Complete Real Data Experiments Results}).
\end{itemize}

Throughout this supplementary material, references to the main paper are highlighted in \textcolor{orange}{orange}, including equation references (e.g., \textcolor{orange}{Eq.}) and section references (e.g., \textcolor{orange}{Sec.}).

\section{Algorithmic Summary}
\label{supp:sec:Algorithmic Summary}

Algorithm~\ref{alg:RSSC} summarizes the overall optimization procedure of the proposed rolling shutter self-calibration framework. The three variants, RSSC-TE, RSSC-CE, and RSSC-DP, share a common iterative optimization structure and differ primarily in the construction of their reprojection residuals. At each iteration, residuals are evaluated according to the selected formulation, accumulated into a global least-squares problem, and used to update the state variables until convergence.

\begin{algorithm}
\caption{Proposed Rolling Shutter Self-Calibration (RSSC)}
\label{alg:RSSC}
\begin{algorithmic}[1]
\renewcommand{\baselinestretch}{1}\selectfont 
\REQUIRE 
    Initial state vector $\boldsymbol{\theta} = \{\mathbf{K}, \{\mathbf{T}_i\}, \{\mathbf{P}^j\}, \gamma$\}.
    Observed RS point measurements $\mathbf{m}_i^j$;
    Method selection $mode \in \{\text{TE}, \text{CE}, \text{DP}\}$;
    Weighting factor $\lambda$ (if $mode = \text{DP}$);
    Maximum iterations $N_{\max}$ and stopping criteria.
\ENSURE
    Refined state vector $\boldsymbol{\theta}^*$.
\STATE Initialize iteration counter $k \gets 0$
\WHILE{$k < N_{\max}$ and stopping criteria not met}
    \STATE \textbf{Build re-projection error}
    \FOR{each camera $i \in \mathcal{F}$}
        \FOR{each point $j \in \mathcal{P}_i$}
            \STATE Compute row timestamp $t(\gamma, v_i^j)$ via \textcolor{orange}{Eq.~(1)}
            \STATE \textbf{Select projection model based on $mode$:}
            \STATE \hspace{1em} \textbf{case} $mode = \text{TE}$ \textbf{:} \hfill (\textcolor{orange}{Sec.~4.1})
            \STATE \hspace{2em} Evaluate row-wise pose $\mathbf{T}(t(\gamma, v_i^j))$ via \textcolor{orange}{Eq.~(5)}
            \STATE \hspace{2em} Compute reprojection error ${}^{\text{TE}}\mathbf{e}_i^j$ via \textcolor{orange}{Eq.~(7)}
            \STATE \hspace{1em} \textbf{case} $mode = \text{CE}$ \textbf{:} \hfill (\textcolor{orange}{Sec.~4.2})
            \STATE \hspace{2em} Rectify RS observation to ${}^{\text{GS}}\mathbf{m}_i^j(t_i)$ via \textcolor{orange}{Eq.~(8)} or \textcolor{orange}{(10)}
            \STATE \hspace{2em} Compute reprojection error ${}^{\text{CE}}\mathbf{e}_i^j$ via \textcolor{orange}{Eq.~(15)}
            \STATE \hspace{1em} \textbf{case} $mode = \text{DP}$ \textbf{:} \hfill (\textcolor{orange}{Sec.~4.3})
            \STATE \hspace{2em} Evaluate RS pose $\mathbf{T}(t(\gamma, v_i^j))$ via \textcolor{orange}{Eq.~(5)}
            \STATE \hspace{2em} Compute RS reprojection error ${}^{\text{TE}}\mathbf{e}_i^j$ via \textcolor{orange}{Eq.~(17)}
            \STATE \hspace{2em} Rectify RS observation to ${}^{\text{GS}}\mathbf{m}_i^j(t_i)$ via \textcolor{orange}{Eq.~(8)} or \textcolor{orange}{(10)}
            \STATE \hspace{2em} Evaluate GS pose $\mathbf{T}(t_i)$ via \textcolor{orange}{Eq.~(16)}
            \STATE \hspace{2em} Compute GS reprojection error ${}^{\text{CE}}\mathbf{e}_i^j$ via \textcolor{orange}{Eq.~(18)}
            \STATE \hspace{2em} Form weighted residual $\mathbf{e}_i^j = [{}^{\text{TE}}\mathbf{e}_i^j;\ \lambda \cdot {}^{\text{CE}}\mathbf{e}_i^j]$
            \STATE \textbf{end}
            \STATE Stack residuals into $\hat{\mathbf{e}}$
            \STATE Build Jacobian $\mathbf{J}_i^j$ 
            \STATE Connect $\mathbf{J}_i^j$ to global $\mathbf{J}$
        \ENDFOR
    \ENDFOR
    \STATE \textbf{Optimization step}
    \STATE Solve normal equation $\mathbf{J}^\top \mathbf{J} \boldsymbol{\delta} = -\mathbf{J}^\top \hat{\mathbf{e}}$
    \STATE Update state vector $\boldsymbol{\theta} \gets \boldsymbol{\theta} + \boldsymbol{\delta}$
    \STATE $k \gets k + 1$
\ENDWHILE
\STATE \RETURN $\boldsymbol{\theta}^*$
\end{algorithmic}
\end{algorithm}

\section{Jacobian Derivation}
\label{supp:sec:Jacobian Derivation for RSSC}
This section presents the derivation of the Jacobian matrices for the two RS self-calibration methods proposed in this work, namely RSSC-TE and RSSC-CE. Since the combined method RSSC-DP essentially integrates these two approaches, its Jacobian can be readily constructed from the corresponding components. For brevity, we omit its explicit derivation.

\subsection{Jacobian Derivation for RSSC-TE}
\label{supp:sec:Jacobian Derivation for RSSC-TE}
This section derives the analytical Jacobians for the proposed RSSC-TE method.

\subsubsection{Residual formulation.}
\label{supp:sec:residual-formulation}
For a 3D point $\mathbf{P}^j=[X^j,Y^j,Z^j]^{\top}$ observed in the $i$-th RS image at pixel $\mathbf{m}_i^j=[u_i^j,v_i^j]^{\top}$, the reprojection residual is defined as:
\begin{equation}
{}^{\text{TE}}\mathbf{e}_i^j = \mathbf{m}_i^j - \pi\Big(\mathbf{K}\mathbf{T}(t_i^j)\mathbf{P}^j\Big),
\label{supp:equ:residual}
\end{equation}
where $\pi(\cdot)$ denotes the perspective projection function, and $\mathbf{K}(f_x,f_y,c_x,c_y)$ denotes the camera intrinsic matrix.

The observation timestamp depends on the readout time ratio $\gamma$ and is given by:
\begin{equation}
t_i^j = \left(i + \gamma\frac{v_i^j}{H}\right)T_f,
\end{equation}
where $H$ is the image height, and $T_f$ denotes the total time interval between two consecutive frames.

The camera pose at time $t_i^j$ is interpolated from the cumulative B-spline~\cite{lovegrove2013spline} trajectory:
\begin{equation}
\mathbf{T}(t_i^j) = \mathbf{T}_{i-1} \prod_{k=1}^{3} \exp\Big(\mathbf{B}_k(s)\boldsymbol{\Omega}_{i-1+k}\Big),
\label{supp:equ:spline_pose}
\end{equation}
where $s = \gamma v_i^j / H$ is the normalized time offset within the current knot interval, and the relative Lie algebra increments are:
\begin{equation}
\boldsymbol{\Omega}_i = \log\bigl( \mathbf{T}_{i-1}^{-1} \mathbf{T}_{i} \bigr) \in \mathfrak{se}(3).
\end{equation}

The cumulative B-spline basis functions $\mathbf{B}(s) \in \mathbb{R}^4$ can be written in matrix form as:
\begin{equation}\label{equ:B-spline-matrix}
\mathbf{B}(s)
=
\frac{1}{6}
\begin{pmatrix}
6 & 0 & 0 & 0 \\
5 & 3 & -3 & 1 \\
1 & 3 & 3 & -2 \\
0 & 0 & 0 & 1
\end{pmatrix}
\begin{pmatrix}
1 \\
s \\
s^2 \\
s^3
\end{pmatrix},
\end{equation}
where $\mathbf{B}_k(s)$ denotes the $k$-th element of $\mathbf{B}(s)$, with indexing starting from $0$. 

Since the interpolated pose depends on four consecutive control poses, the Jacobian of a single residual has the following block structure:
\begin{equation}
{\changefontsize{9pt}
{}^{\text{TE}}\mathbf{J}_i^j = \left[ 
\frac{\partial {}^{\text{TE}}\mathbf{e}_i^j}{\partial \boldsymbol{\xi}_{i-1}},\; 
\frac{\partial {}^{\text{TE}}\mathbf{e}_i^j}{\partial \boldsymbol{\xi}_{i}},\; 
\frac{\partial {}^{\text{TE}}\mathbf{e}_i^j}{\partial \boldsymbol{\xi}_{i+1}},\; 
\frac{\partial {}^{\text{TE}}\mathbf{e}_i^j}{\partial \boldsymbol{\xi}_{i+2}},\; 
\frac{\partial {}^{\text{TE}}\mathbf{e}_i^j}{\partial \mathbf{P}^j},\; 
\frac{\partial {}^{\text{TE}}\mathbf{e}_i^j}{\partial \mathbf{K}},\; 
\frac{\partial {}^{\text{TE}}\mathbf{e}_i^j}{\partial \gamma} 
\right]_{2\times32},
\label{supp:equ:jacobian_structure}
}
\end{equation}
where $\boldsymbol{\xi}_i = \log(\mathbf{T}_i) \in \mathfrak{se}(3)$ denotes the Lie algebra representation of the $i$-th control pose of the B-spline trajectory.

\subsubsection{Jacobian with respect to the interpolated pose.}
\label{supp:sec:projection-jacobian}

Let the 3D point $\mathbf{P}^j$ be transformed into the camera frame as:
\begin{equation}
\mathbf{P}_c^j = \mathbf{T}(t_i^j)\mathbf{P}^j = [X_c^j,\;Y_c^j,\;Z_c^j]^{\top}.
\end{equation}

The perspective projection with camera intrinsics $\mathbf{K}(f_x,f_y,c_x,c_y)$ gives
\begin{equation}
u_i^j = f_x \frac{X_c^j}{Z_c^j} + c_x,\qquad 
v_i^j = f_y \frac{Y_c^j}{Z_c^j} + c_y.
\end{equation}

The Jacobian of the reprojection error with respect to  interpolated pose $\boldsymbol{\xi}_i^j$ is given by
\begin{equation}
{\changefontsize{6.5pt}
\mathbf{J_{\pi}}_i^j = \frac{\partial {}^{\text{TE}}\mathbf{e}_i^j} {\partial \boldsymbol{\xi}_i^j} =
\begin{bmatrix} 
-\dfrac{f_x X_c^j Y_c^j}{(Z_c^j)^2} & \left(f_x + \dfrac{f_x (X_c^j)^2}{(Z_c^j)^2}\right) & -\dfrac{f_x Y_c^j}{Z_c^j} & \dfrac{f_x}{Z_c^j} & 0 & -\dfrac{f_x X_c^j}{(Z_c^j)^2} \\[6pt] 
-\left(f_y + \dfrac{f_y (Y_c^j)^2}{(Z_c^j)^2}\right) & \dfrac{f_y X_c^j Y_c^j}{(Z_c^j)^2} & \dfrac{f_y X_c^j}{Z_c^j} & 0 & \dfrac{f_y}{Z_c^j} & -\dfrac{f_y Y_c^j}{(Z_c^j)^2} 
\end{bmatrix}_{2\times6}.
\label{supp:equ:jac_proj}
}
\end{equation}

\subsubsection{Jacobian with respect to control poses.}
\label{supp:sec:jacobian-control-poses}

Since the interpolated pose $\mathbf{T}(t_i^j)$ depends on four consecutive control poses
$\{\mathbf{T}_{i-1},\mathbf{T}_i,\mathbf{T}_{i+1},\mathbf{T}_{i+2}\}$,
the Jacobian of a single residual with respect to a control pose
$\boldsymbol{\xi}_\ell$ ($\ell\in\{i-1,i,i+1,i+2\}$) is
\begin{equation}
\frac{\partial {}^{\text{TE}}\mathbf{e}_i^j}{\partial \boldsymbol{\xi}_\ell}
=
\mathbf{J_{\pi}}_i^j
\frac{\partial \boldsymbol{\xi}_i^j}{\partial \boldsymbol{\xi}_\ell}.
\label{supp:eq:chain_pose_refined}
\end{equation}

For notational convenience, Eq.~(\ref{supp:equ:spline_pose}) is rewritten in the following factorized form:
\begin{equation}
\mathbf{T}(t_i^j)
=
\mathbf{A}_0
\mathbf{A}_1
\mathbf{A}_2
\mathbf{A}_3 ,
\end{equation}
where
\begin{equation}
\begin{aligned}
\mathbf{A}_k &= \mathbf{T}_{i-1}, 
&\qquad k &= 0, \\
\mathbf{A}_k &= 
\exp\!\left(
\mathbf{B}_k(s)\boldsymbol{\Omega}_{i-1+k}
\right), 
&\qquad k &= 1,2,3 .
\end{aligned}
\end{equation}

We define $\mathbf{a}_k = \log(\mathbf{A}_k) \in \mathfrak{se}(3)$ as the corresponding Lie algebra elements.
Therefore, Eq.~(\ref{supp:eq:chain_pose_refined}) can be further expressed as:
\begin{equation} 
\frac{\partial {}^{\text{TE}}\mathbf{e}_i^j}{\partial \boldsymbol{\xi}_\ell} 
= 
\mathbf{J_{\pi}}_i^j 
\left(
\frac{\partial \boldsymbol{\xi}_i^j}{\partial \mathbf{a}_{\ell-i+1}}
\frac{\partial \mathbf{a}_{\ell-i+1}}{\partial \boldsymbol{\xi}_\ell}
+
\frac{\partial \boldsymbol{\xi}_i^j}{\partial \mathbf{a}_{\ell-i+2}}
\frac{\partial \mathbf{a}_{\ell-i+2}}{\partial \boldsymbol{\xi}_\ell}
\right),
\label{supp:eq:chain_pose_refined_all}
\end{equation} 
where $\ell \in \{i-1,i,i+1,i+2\}$.

To represent the cumulative transformations preceding each exponential term compactly, we introduce the auxiliary variables:
\begin{equation}
\begin{aligned}
\mathbf{G}_0 &= \mathbf{I}, 
&\qquad k &= 0, \\
\mathbf{G}_k &= \mathbf{G}_{k-1}\mathbf{A}_{k-1}, 
&\qquad k &= 1,2,3 .
\end{aligned}
\end{equation}

With this definition, the derivatives of $\boldsymbol{\xi}_i^j$ with respect to the intermediate variables $\mathbf{a}_k$ take the following compact form:
\begin{equation}
\begin{aligned}
\frac{\partial \boldsymbol{\xi}_i^j}{\partial \mathbf{a}_k} &= \mathbf{I}, 
&\qquad k &= 0, \\
\frac{\partial \boldsymbol{\xi}_i^j}{\partial \mathbf{a}_k}
&=
\mathrm{Ad}_{\mathbf{G}_{k}}
\mathbf{J}_l(\mathbf{a}_k),
&\qquad k &= 1,2,3 .
\end{aligned}
\label{supp:eq:xi_a}
\end{equation}

Next, we compute the derivatives of the intermediate Lie algebra variables $\mathbf{a}_k$ with respect to the control pose perturbations $\boldsymbol{\xi}_\ell$:
\begin{equation}
\begin{aligned}
\frac{\partial a_k}{\partial \boldsymbol{\xi}_{i+k-2}} &=  -\mathbf{B}_k(s) \mathbf{J}_l^{-1}(\boldsymbol{\Omega}_{i+k-1})
\mathrm{Ad}_{\mathbf{T}_{i+k-2}^{-1}}, 
&\qquad k &= 1,2,3 \\
\frac{\partial a_k}{\partial \boldsymbol{\xi}_{i+k-1}}
&=\mathbf{I},
&\qquad k &= 0 ,
\\
\frac{\partial a_k}{\partial \boldsymbol{\xi}_{i+k-1}}
&=-\frac{\partial a_k}{\partial \boldsymbol{\xi}_{i+k-2}},
&\qquad k &= 1,2,3 .
\end{aligned}
\label{supp:eq:a_xi}
\end{equation}

Finally, substituting Eq.~(\ref{supp:eq:xi_a}) and Eq.~(\ref{supp:eq:a_xi}) into Eq.~(\ref{supp:eq:chain_pose_refined_all}) gives the Jacobians of the residual with respect to the four control poses:
\begin{equation}
\footnotesize
\begin{aligned}[b]
\left\{
\begin{aligned}
\frac{\partial {}^{\text{TE}}\mathbf{e}_i^j}{\partial \boldsymbol{\xi}_{i-1}} 
&= \mathbf{J_{\pi}}_i^j
\left(
\mathbf{I}
-
\mathbf{B}_1(s)
\mathrm{Ad}_{\mathbf{G}_1}
\mathbf{J}_l(\mathbf{a}_1)
\mathbf{J}_l^{-1}(\boldsymbol{\Omega}_i)
\mathrm{Ad}_{\mathbf{T}_{i-1}^{-1}}
\right),
\\
\frac{\partial {}^{\text{TE}}\mathbf{e}_i^j}{\partial \boldsymbol{\xi}_{i}} 
&= \mathbf{J_{\pi}}_i^j
\left(
\mathbf{B}_1(s)
\mathrm{Ad}_{\mathbf{G}_1}
\mathbf{J}_l(\mathbf{a}_1)
\mathbf{J}_l^{-1}(\boldsymbol{\Omega}_i)
\mathrm{Ad}_{\mathbf{T}_{i-1}^{-1}}
+
\mathbf{B}_2(s)
\mathrm{Ad}_{\mathbf{G}_2}
\mathbf{J}_l(\mathbf{a}_2)
\mathbf{J}_l^{-1}(\boldsymbol{\Omega}_{i+1})
\mathrm{Ad}_{\mathbf{T}_{i}^{-1}}
\right),
\\
\frac{\partial {}^{\text{TE}}\mathbf{e}_i^j}{\partial \boldsymbol{\xi}_{i+1}} 
&= \mathbf{J_{\pi}}_i^j
\left(
\mathbf{B}_2(s)
\mathrm{Ad}_{\mathbf{G}_2}
\mathbf{J}_l(\mathbf{a}_2)
\mathbf{J}_l^{-1}(\boldsymbol{\Omega}_{i+1})
\mathrm{Ad}_{\mathbf{T}_{i}^{-1}}
+
\mathbf{B}_3(s)
\mathrm{Ad}_{\mathbf{G}_3}
\mathbf{J}_l(\mathbf{a}_3)
\mathbf{J}_l^{-1}(\boldsymbol{\Omega}_{i+2})
\mathrm{Ad}_{\mathbf{T}_{i+1}^{-1}}
\right),
\\
\frac{\partial {}^{\text{TE}}\mathbf{e}_i^j}{\partial \boldsymbol{\xi}_{i+2}} 
&= \mathbf{J_{\pi}}_i^j
\,
\left(
\mathbf{B}_3(s)
\mathrm{Ad}_{\mathbf{G}_3}
\mathbf{J}_l(\mathbf{a}_3)
\mathbf{J}_l^{-1}(\boldsymbol{\Omega}_{i+2})
\mathrm{Ad}_{\mathbf{T}_{i+1}^{-1}}
\right),
\end{aligned}
\right.
\end{aligned}
\end{equation}
which is a $2\times 24$ Jacobian corresponding to the final block in Eq.~(\ref{supp:equ:jacobian_structure}).

\subsubsection{Jacobian with respect to 3D point.}
\label{supp:sec:jacobian-3d-points}
The Jacobian of the residual with respect to the 3D point $\mathbf{P}^j$ is
\begin{equation}
\frac{\partial {}^{\text{TE}}\mathbf{e}_i^j}{\partial \mathbf{P}^j} = 
\frac{\partial {}^{\text{TE}}\mathbf{e}_i^j}{\partial \mathbf{P}_c^j} \; 
\frac{\partial \mathbf{P}_c^j}{\partial \mathbf{P}^j}.
\label{supp:eq:jac_chain_te}
\end{equation}

The derivative of the residual with respect to the point in the camera frame follows from the projection model:
\begin{equation}
\frac{\partial {}^{\text{TE}}\mathbf{e}_i^j}{\partial \mathbf{P}_c^j} =
- \begin{bmatrix}
\dfrac{f_x}{Z_c^j} & 0 & -\dfrac{f_x X_c^j}{(Z_c^j)^2} \\[3pt]
0 & \dfrac{f_y}{Z_c^j} & -\dfrac{f_y Y_c^j}{(Z_c^j)^2}
\end{bmatrix}_{2\times3}.
\label{supp:eq:jac_proj}
\end{equation}

The derivative of the camera-frame point with respect to the world 3D point is given by the rotation component of the interpolated pose:
\begin{equation}
\frac{\partial \mathbf{P}_c^j}{\partial \mathbf{P}^j} = \mathbf{R}(t_i^j),
\label{supp:eq:jac_rot_te}
\end{equation}
where $\mathbf{R}(t_i^j)$ is the rotation part of the interpolated pose $\mathbf{T}(t_i^j)$.

Substituting Eq.~(\ref{supp:eq:jac_proj}) and Eq.~(\ref{supp:eq:jac_rot_te}) into Eq.~(\ref{supp:eq:jac_chain_te}) gives the Jacobian of the residual with respect to the 3D point $\mathbf{P}^j$, which is a $2\times 3$ block corresponding to the point parameters in Eq.~(\ref{supp:equ:jacobian_structure}).

\subsubsection{Jacobian with respect to camera intrinsics.}
\label{supp:sec:jacobian-intrinsics}

The Jacobian of the residual with respect to the camera intrinsic parameters 
$\mathbf{K} = (f_x, f_y, c_x, c_y)$ is
\begin{equation}
\frac{\partial {}^{\text{TE}}\mathbf{e}_i^j}{\partial \mathbf{K}} =
- \begin{bmatrix}
\dfrac{X_c^j}{Z_c^j} & 0 & 1 & 0\\[3pt]
0 & \dfrac{Y_c^j}{Z_c^j} & 0 & 1
\end{bmatrix}_{2\times4},
\label{supp:eq:jac_intrinsic}
\end{equation}
which is a $2\times 4$ Jacobian corresponding to the final block in Eq.~(\ref{supp:equ:jacobian_structure}).

\subsubsection{Jacobian with respect to readout time ratio.}
\label{supp:sec:jacobian-readout-ratio}

The Jacobian of the residual with respect to the readout ratio $\gamma$ is
\begin{equation}
\frac{\partial {}^{\text{TE}}\mathbf{e}_i^j}{\partial \gamma}
=
\mathbf{J_{\pi}}_i^j
\frac{\partial \boldsymbol{\xi}_i^j}{\partial \gamma}.
\end{equation}

The derivative of $\boldsymbol{\xi}_i^j$ with respect to $\gamma$ can be expressed as a sum over the B-spline basis functions $\mathbf{B}_k(s)$:
\begin{equation}
\frac{\partial \boldsymbol{\xi}_i^j}{\partial \gamma}
=
\sum_{k=1}^{3} 
\frac{\partial \boldsymbol{\xi}_i^j}{\partial \mathbf{B}_k(s)}
\frac{\mathrm{d} \mathbf{B}_k(s)}{\mathrm{d} s}
\frac{\partial s}{\partial \gamma}.
\label{supp:eq:jac_chain}
\end{equation}
The Jacobian of $\boldsymbol{\xi}_i^j$ with respect to each scaled basis $\mathbf{B}_k(s)$ is
\begin{equation}
\frac{\partial \boldsymbol{\xi}_i^j}{\partial \mathbf{B}_k(s)} = 
\mathrm{Ad}_{\mathbf{G}_k} \mathbf{J}_l(\mathbf{a}_k) \boldsymbol{\Omega}_{i-1+k}, 
\qquad k=1,2,3.
\label{supp:eq:jac_rot}
\end{equation}
The derivatives of the basis functions with respect to $s$ are
\begin{equation}
\frac{\mathrm{d} \mathbf{B}_1(s)}{\mathrm{d} s} = \frac{1}{2} (1 - 2s + s^2), \quad
\frac{\mathrm{d} \mathbf{B}_2(s)}{\mathrm{d} s} = \frac{1}{2} (1 + 2s - 2s^2), \quad
\frac{\mathrm{d} \mathbf{B}_3(s)}{\mathrm{d} s} = \frac{1}{2} s^2,
\label{supp:equ:dB_ds_compact}
\end{equation}
and the derivative of $s$ with respect to $\gamma$ is
\begin{equation}
\frac{\partial s}{\partial \gamma} = \frac{v_i^j}{H}.
\label{supp:eq:ds_dgamma}
\end{equation}

Substituting Eq.~(\ref{supp:eq:jac_rot}), Eq.~(\ref{supp:equ:dB_ds_compact}), and Eq.~(\ref{supp:eq:ds_dgamma}) into Eq.~(\ref{supp:eq:jac_chain}) gives the final Jacobian with respect to $\gamma$:
\begin{equation}
\frac{\partial {}^{\text{TE}}\mathbf{e}_i^j}{\partial \gamma}
=
\mathbf{J_{\pi}}_i^j
\left(
\frac{v_i^j}{H}
\sum_{k=1}^{3} 
\frac{\mathrm{d} \mathbf{B}_k(s)}{\mathrm{d} s}
\mathrm{Ad}_{\mathbf{G}_k} \mathbf{J}_l(\mathbf{a}_k) \boldsymbol{\Omega}_{i-1+k}
\right),
\end{equation}
which is a $2\times 1$ Jacobian corresponding to the final block in Eq.~(\ref{supp:equ:jacobian_structure}).

\subsection{Jacobian Derivation for RSSC-CE}
\label{supp:sec:Jacobian Derivation for RSSC-CE}

This section derives the analytical Jacobians for the proposed RSSC-CE method.

\subsubsection{Residual formulation.}
\label{supp:sec:ce_residual_formulation}

For a 3D point $\mathbf{P}^j$ observed in frame $i$, the residual is defined as:
\begin{equation}
{}^{\text{CE}}\mathbf{e}_i^j
=
{}^{\text{GS}}\mathbf{m}_{i}^{j}(t_i)
-
\pi\!\left(
\mathbf{K}\,{}^{\text{GS}}\mathbf{T}_i\,\mathbf{P}^j
\right),
\label{supp:eq:ce_residual}
\end{equation}
where ${}^{\text{GS}}\mathbf{m}_{i}^{j}(t_i)$ is the rectified 2D observation at the GS timestamp $t_i$, ${}^{\text{GS}}\mathbf{T}_i$ is the corresponding camera pose, and ${}^{\text{GS}}\boldsymbol{\xi}_i = \log({}^{\text{GS}}\mathbf{T}_i) \in \mathfrak{se}(3)$ is its Lie algebra representation.

The Jacobian of the residual with respect to the parameters is structured as:
\begin{equation}
\mathbf{J}_i^j
=
\left[
\frac{\partial {}^{\text{CE}}\mathbf{e}_i^j}{\partial {}^{\text{GS}}\boldsymbol{\xi}_{i}},
\;
\frac{\partial {}^{\text{CE}}\mathbf{e}_i^j}{\partial \mathbf{P}^j},
\;
\frac{\partial {}^{\text{CE}}\mathbf{e}_i^j}{\partial \mathbf{K}},
\;
\frac{\partial {}^{\text{CE}}\mathbf{e}_i^j}{\partial \gamma}
\right]_{2\times14}.
\label{supp:eq:ce_jac_structure}
\end{equation}

For both RSSC-CEQ and RSSC-CEH, all Jacobian blocks are identical, with the exception of the derivative with respect to the readout ratio $\gamma$.

\subsubsection{Jacobian with respect to camera pose.}
\label{supp:sec:ce_jac_pose}

Let the 3D point $\mathbf{P}^j$ in GS camera coordinates be:
\begin{equation}
{}^{\text{GS}}\mathbf{P}_c^j
=
{}^{\text{GS}}\mathbf{T}_i\,\mathbf{P}^j
=
[{}^{\text{GS}}X_c^j,\;{}^{\text{GS}}Y_c^j,\;{}^{\text{GS}}Z_c^j]^{\top}.
\end{equation}

The residual Jacobian with respect to the GS camera pose can be obtained by substituting ${}^{\text{GS}}\mathbf{P}_c^j$ and the camera intrinsics $\mathbf{K}$ into the projection Jacobian derived in Eq.~(\ref{supp:equ:jac_proj}):
\begin{equation}
\frac{\partial {}^{\text{CE}}\mathbf{e}_i^j}
{\partial {}^{\text{GS}}\boldsymbol{\xi}_i}
=
{}^{\text{GS}}\mathbf{J_{\pi}}_i^j ,
\end{equation}
which is a $2\times6$ Jacobian corresponding to the first block in Eq.~(\ref{supp:eq:ce_jac_structure}).

\subsubsection{Jacobian with respect to 3D point.}
\label{supp:sec:ce_jac_3d}

The Jacobian of the residual with respect to the 3D point $\mathbf{P}^j$ is:
\begin{equation}
\frac{\partial {}^{\text{CE}}\mathbf{e}_i^j}{\partial \mathbf{P}^j} =
\frac{\partial {}^{\text{CE}}\mathbf{e}_i^j}{\partial {}^{\text{GS}}\mathbf{P}_c^j} \;
\frac{\partial {}^{\text{GS}}\mathbf{P}_c^j}{\partial \mathbf{P}^j},
\label{supp:eq:jac_chain_ce}
\end{equation}

The derivative of the residual with respect to the point in the camera frame follows from the projection model:
\begin{equation}
\frac{\partial {}^{\text{CE}}\mathbf{e}_i^j}{\partial {}^{\text{GS}}\mathbf{P}_c^j} =
- \begin{bmatrix}
\dfrac{f_x}{{}^{\text{GS}}Z_c^j} & 0 & -\dfrac{f_x {}^{\text{GS}}X_c^j}{({}^{\text{GS}}Z_c^j)^2} \\[2mm]
0 & \dfrac{f_y}{{}^{\text{GS}}Z_c^j} & -\dfrac{f_y {}^{\text{GS}}Y_c^j}{({}^{\text{GS}}Z_c^j)^2}
\end{bmatrix}_{2\times3}.
\label{supp:eq:jac_proj_ce}
\end{equation}

The derivative of the camera-frame point with respect to the world 3D point is the rotation part of the GS pose:
\begin{equation}
\frac{\partial {}^{\text{GS}}\mathbf{P}_c^j}{\partial \mathbf{P}^j} = {}^{\text{GS}}\mathbf{R}_i,
\label{supp:eq:jac_rot_ce}
\end{equation}
where ${}^{\text{GS}}\mathbf{R}_i$ is the rotation component of ${}^{\text{GS}}\mathbf{T}_i$.

Substituting Eq.~(\ref{supp:eq:jac_proj_ce}) and Eq.~(\ref{supp:eq:jac_rot_ce}) into Eq.~(\ref{supp:eq:jac_chain_ce}) gives the Jacobian of the residual with respect to the 3D point $\mathbf{P}^j$, which is a $2\times 3$ block corresponding to the point parameters in Eq.~(\ref{supp:eq:ce_jac_structure}).

\subsubsection{Jacobian with respect to camera intrinsics.}
\label{supp:sec:ce_jac_intrinsics}

The Jacobian of the residual with respect to the camera intrinsics
$\mathbf{K} = (f_x, f_y, c_x, c_y)$ is
\begin{equation}
\frac{\partial {}^{\text{CE}}\mathbf{e}_i^j}{\partial \mathbf{K}} =
- \begin{bmatrix}
\dfrac{{}^{\text{GS}}X_c^j}{{}^{\text{GS}}Z_c^j} & 0 & 1 & 0\\[2mm]
0 & \dfrac{{}^{\text{GS}}Y_c^j}{{}^{\text{GS}}Z_c^j} & 0 & 1
\end{bmatrix}_{2\times4},
\end{equation}
which is a $2\times4$ Jacobian corresponding to the intrinsics block in Eq.~(\ref{supp:eq:ce_jac_structure}).

\subsubsection{Jacobian with respect to readout time ratio.}
\label{supp:sec:ce_jac_gamma}

The readout ratio $\gamma$ affects the residual through the rectified 2D pixel trajectory ${}^{\text{GS}}\mathbf{m}_i^j(t_i)$. 
The residual Jacobian with respect to $\gamma$ is 
\begin{equation}
\frac{\partial {}^{\text{CE}}\mathbf{e}_i^j}{\partial \gamma} =
- \frac{\partial {}^{\text{GS}}\mathbf{m}_i^j(t_i)}{\partial \gamma}.
\label{supp:eq:ce_jac_gamma}
\end{equation}

The expression of $\partial {}^{\text{GS}}\mathbf{m}_i^j / \partial \gamma$ depends on the temporal interpolation used in the RS self-calibration method.  

In RSSC-CEQ, the rectified point ${}^{\text{GS}}\mathbf{m}_i^j(t_i)$ is computed via Lagrange interpolation of the three temporally adjacent observations from frames $i-1$, $i$, and $i+1$.
\begin{equation}
{}^{\text{GS}}\mathbf{m}_i^j(t_i) =
L_0(t_i)\mathbf{m}_{i-1}^{j} +
L_1(t_i)\mathbf{m}_{i}^{j} +
L_2(t_i)\mathbf{m}_{i+1}^{j},
\end{equation}

where the Lagrange basis functions are:
\begin{equation}
\begin{cases}
L_0(t_{i}) = \frac{(t_i - t(\gamma, v_{i}^{j}))(t_i - t(\gamma, v_{i+1}^{j}))}{(t(\gamma, v_{i-1}^{j}) - t(\gamma, v_{i}^{j}))(t(\gamma, v_{i-1}^{j}) - t(\gamma, v_{i+1}^{j}))},\\
L_1(t_{i}) = \frac{(t_i - t(\gamma, v_{i-1}^{j}))(t_i - t(\gamma, v_{i+1}^{j}))}{(t(\gamma, v_{i}^{j}) - t(\gamma, v_{i-1}^{j}))(t(\gamma, v_{i}^{j}) - t(\gamma, v_{i+1}^{j}))},\\
L_2(t_{i}) = \frac{(t_i - t(\gamma, v_{i-1}^{j}))(t_i - t(\gamma, v_{i}^{j}))}{(t(\gamma, v_{i+1}^{j}) - t(\gamma, v_{i-1}^{j}))(t(\gamma, v_{i+1}^{j}) - t(\gamma, v_{i}^{j}))}.
\end{cases}
\end{equation}

For notational simplicity, we denote the timestamps of the three consecutive observations as:
\begin{equation}
t_g = t(\gamma, v_{g+i-1}^{j}) = \Bigl(g+i-1+\gamma\frac{v_{g+i-1}^j}{H}\Bigr)T_f,\quad g = 0,1,2.
\end{equation}
So, the Jacobian of the rectified observation with respect to $\gamma$ is
\begin{equation}
\frac{\partial {}^{\text{GS}}\mathbf{m}_i^j}{\partial \gamma} =
\mathbf{m}_{i-1}^{j} \frac{\partial L_0(t_i)}{\partial \gamma} +
\mathbf{m}_{i}^{j} \frac{\partial L_1(t_i)}{\partial \gamma} +
\mathbf{m}_{i+1}^{j} \frac{\partial L_2(t_i)}{\partial \gamma}.
\label{supp:eq:dm-L-gamma}
\end{equation}

The derivatives with respect to $\gamma$ are
\begin{equation}
\begin{cases}
\frac{\partial L_0(t_i)}{\partial \gamma} = -\frac{T_f}{H}\frac{(t_i-t_2)v_i^j + (t_i-t_1)v_{i+1}^j}{(t_0-t_1)(t_0-t_2)},\\
\frac{\partial L_1(t_i)}{\partial \gamma} = -\frac{T_f}{H}\frac{(t_i-t_2)v_{i-1}^j + (t_i-t_0)v_{i+1}^j}{(t_1-t_0)(t_1-t_2)},\\
\frac{\partial L_2(t_i)}{\partial \gamma} = -\frac{T_f}{H}\frac{(t_i-t_1)v_{i-1}^j + (t_i-t_0)v_{i}^j}{(t_2-t_0)(t_2-t_1)},
\label{supp:eq:L-gamma}
\end{cases}
\end{equation}

Finally, substituting Eq.~(\ref{supp:eq:L-gamma}) into Eq.~(\ref{supp:eq:dm-L-gamma}) yields the complete Jacobian of the rectified observation with respect to $\gamma$.

\paragraph{Cubic Hermite Interpolation.}
In RSSC-CEH, the rectified point ${}^{\text{GS}}\mathbf{m}_i^j(t_i)$ is computed via cubic Hermite interpolation of four consecutive observations from frames $i-1$ to $i+2$:
\begin{equation}
{}^{\text{GS}}\mathbf{m}_i^j(t_i) = 
h_{00}(\tau)\mathbf{m}_{i}^{j} + 
h_{10}(\tau)\Delta t\,\mathbf{d}_i + 
h_{01}(\tau)\mathbf{m}_{i+1}^{j} + 
h_{11}(\tau)\Delta t\,\mathbf{d}_{i+1},
\end{equation}
where
\begin{equation}
\tau = \frac{t_i - t_1}{\Delta t},\quad
\Delta t = t(\gamma, v_{i+1}^{j})-t(\gamma, v_{i}^{j}),
\end{equation}
and the Hermite basis functions are
\begin{equation}
\begin{cases}
\begin{aligned}
h_{00}(\tau) &= 2\tau^3 - 3\tau^2 + 1,\\
h_{10}(\tau) &= \tau^3 - 2\tau^2 + \tau,\\
h_{01}(\tau) &= -2\tau^3 + 3\tau^2,\\
h_{11}(\tau) &= \tau^3 - \tau^2.
\end{aligned}
\end{cases}
\end{equation}
The tangent vectors are approximated by central differences:
\begin{equation}
\mathbf{d}_{i} = 
\frac{\mathbf{m}_{i+1}^{j} - \mathbf{m}_{i-1}^{j}}{t_2 - t_0},\quad
\mathbf{d}_{i+1} = 
\frac{\mathbf{m}_{i+2}^{j} - \mathbf{m}_{i}^{j}}{t_3 - t_1}.
\end{equation}
The timestamps of the four consecutive observations are simplified as:
\begin{equation}
t_g = t(\gamma, v_{g+i-1}^{j}) = (g+i-1+\gamma\frac{v_{g+i-1}^j}{H})T_f,\quad g=0,1,2,3.
\label{supp:eq:t-gamma}
\end{equation}
Therefore, the rectified observation Jacobian with respect to $\gamma$ is
\begin{equation}
\begin{aligned}
\frac{\partial {}^{\text{GS}}\mathbf{m}_i^j}{\partial \gamma}
=&
\mathbf{m}_i^j \frac{\partial h_{00}}{\partial \gamma} + 
\mathbf{m}_{i+1}^j \frac{\partial h_{01}}{\partial \gamma} +
\Delta t \Big(\mathbf{d}_i \frac{\partial h_{10}}{\partial \gamma} + \mathbf{d}_{i+1} \frac{\partial h_{11}}{\partial \gamma}\Big) \\
&+
(h_{10}\mathbf{d}_i + h_{11}\mathbf{d}_{i+1}) \frac{\partial \Delta t}{\partial \gamma} +
h_{10}\Delta t \frac{\partial \mathbf{d}_i}{\partial \gamma} +
h_{11}\Delta t \frac{\partial \mathbf{d}_{i+1}}{\partial \gamma}.
\label{supp:eq:dm-gamma}
\end{aligned}
\end{equation}

The derivatives of the Hermite basis functions with respect to the readout time ratio $\gamma$ are
\begin{equation}
\frac{\partial h_{ab}}{\partial \gamma} = \frac{d h_{ab}}{d\tau}\frac{\partial \tau}{\partial \gamma}, \quad h_{ab}\in \{h_{00},h_{10},h_{01},h_{11}\},
\label{supp:eq:dh-gamma}
\end{equation}
with
\begin{equation}
\begin{cases}
\begin{aligned}
\frac{d h_{00}}{d\tau} &= 6\tau^2 - 6\tau,\\
\frac{d h_{10}}{d\tau} &= 3\tau^2 - 4\tau + 1,\\
\frac{d h_{01}}{d\tau} &= -6\tau^2 + 6\tau,\\
\frac{d h_{11}}{d\tau} &= 3\tau^2 - 2\tau,
\end{aligned}
\end{cases}
\label{supp:eq:dh-dtau}
\end{equation}
and
\begin{equation}
\frac{\partial \tau}{\partial \gamma} = \frac{-(\partial t_1/\partial \gamma)(t_2-t_1) - (t_i-t_1)(\partial t_2/\partial \gamma-\partial t_1/\partial \gamma)}{(t_2-t_1)^2}.
\label{supp:eq:tau-dt-gamma}
\end{equation}
The derivative of the $\Delta t$ with respect to the readout time ratio $\gamma$ is
\begin{equation}
\frac{\partial \Delta t}{\partial \gamma} = \frac{\partial t_2}{\partial \gamma}-\frac{\partial t_1}{\partial \gamma}, \quad \frac{\partial t_g}{\partial \gamma} = \frac{v_{g+i-1}^j}{H}T_f,
\label{supp:eq:dtg-dgamma}
\end{equation}
and the derivatives of the tangent vectors with respect to the readout time ratio $\gamma$ are
\begin{equation}
\small
\frac{\partial \mathbf{d}_i}{\partial \gamma} = -\frac{\mathbf{m}_{i+1}^j-\mathbf{m}_{i-1}^j}{(t_2-t_0)^2}\Big(\frac{\partial t_2}{\partial \gamma}-\frac{\partial t_0}{\partial \gamma}\Big),\quad
\frac{\partial \mathbf{d}_{i+1}}{\partial \gamma} = -\frac{\mathbf{m}_{i+2}^j-\mathbf{m}_{i}^{j}}{(t_3-t_1)^2}\Big(\frac{\partial t_3}{\partial \gamma}-\frac{\partial t_1}{\partial \gamma}\Big).
\label{supp:eq:dd-gamma}
\end{equation}

Finally, substituting Eq.~(\ref{supp:eq:dh-gamma}), Eq.~(\ref{supp:eq:dtg-dgamma}), and Eq.~(\ref{supp:eq:dd-gamma}) into Eq.~(\ref{supp:eq:dm-gamma}) yields the complete Jacobian of the rectified observation with respect to $\gamma$.

\section{Complete Real Data Experiments Results}
\label{supp:sec:Complete Real Data Experiments Results}

In this section, we present the complete set of results from the real data experiments.

\begin{table*}[t]
\centering
\caption{Self-calibration results on the WHU-RSVI dataset~\cite{cao2020whu}. 
The table reports absolute errors (AE) of the intrinsic parameters and readout time ratio, as well as the absolute trajectory error (ATE), for sequences WHU-RSVI1 and WHU-RSVI2.}
\scalebox{1}[1]{
\begin{tabular}{l l c c c c c c}
\Xhline{1pt}
Dataset & Method & AE($f_x$)& AE($f_y$) & AE($c_x$) & AE($c_y$)  & AE($\gamma$) & ATE \\
\Xhline{1pt}

\multirow{8}{*}{WHU-1} 
& COLMAP~\cite{schoenberger2016sfm} & 5.33 & 14.90 & 5.90 & 10.73 & --- & 0.0470 \\
& DroidCalib~\cite{hagemann2023deep} & 0.37 & 38.85 & 5.52 & 24.13 & --- & 0.0405 \\
& SelfSup-Calib~\cite{fang2022self} & 241.54 & 179.21 & 26.21 & 25.87 & --- & 0.3879 \\
& RSSC-TE & \textbf{0.03} & \underline{0.16} & \underline{0.04} & 1.04 & \underline{0.0583} & 0.0228 \\
& RSSC-CEQ & 0.06 & 4.64 & 0.12 & 1.68 & 0.0845 & 0.0094 \\
& RSSC-CEH & 0.83 & 3.70 & 1.35 & \textbf{0.82} & 0.1657 & 0.0131 \\
& RSSC-DPQ & \textbf{0.03} & \textbf{0.15} & \textbf{0.01} & 1.11 & 0.0539 & \underline{0.0081} \\
& RSSC-DPH & 0.07 & 0.29 & 0.10 & \underline{0.94} & \textbf{0.0510} & \textbf{0.0073} \\

\midrule

\multirow{8}{*}{WHU-2}
& COLMAP~\cite{schoenberger2016sfm} & 0.76 & 7.61 & 2.82 & 5.60 & --- & 0.0435 \\
& DroidCalib~\cite{hagemann2023deep} & 0.74 & 5.81 & 0.73 & 3.46 & --- & 0.0229 \\
& SelfSup-Calib~\cite{fang2022self} & 150.42 & 79.35 & 1.48 & 21.69 & --- & 0.5291 \\
& RSSC-TE & 0.11 & \underline{0.45} & 0.42 & \textbf{0.16} & \textbf{0.0387} & \textbf{0.0112} \\
& RSSC-CEQ & 0.25 & 1.00 & 0.42 & 0.39 & 0.0782 & 0.0168 \\
& RSSC-CEH & 0.50 & 1.15 & \textbf{0.06} & 2.38 & 0.1920 & 0.0344 \\
& RSSC-DPQ & \textbf{0.03} & 0.73 & \underline{0.37} & \underline{0.19} & \underline{0.0427} & \underline{0.0122} \\
& RSSC-DPH & 0.52 & \textbf{0.33} & \underline{0.37} & 2.05 & 0.1380 & 0.0179 \\

\Xhline{1pt}
\end{tabular}
}
\label{table:self-calibration-results1}
\end{table*}

\noindent\textbf{Datasets and compared methods.} 
We conduct self-calibration evaluation on two publicly available RS datasets, WHU-RSVI~\cite{cao2020whu} and TUM-RSVI~\cite{dai2016rolling}. For each sequence, 50 consecutive frames are selected for calibration. Our method uses COLMAP~\cite{schoenberger2016sfm} to obtain an initial reconstruction, with the readout time ratio initialized to 0.5. The comparison methods include \textit{1)} COLMAP~\cite{schoenberger2016sfm}, \textit{2)} DroidCalib~\cite{hagemann2023deep}, and \textit{3)} SelfSup-Calib~\cite{fang2022self}. Since no publicly available methods exist for self-calibration of RS cameras, only approaches designed for GS cameras are included in the comparison.

\noindent\textbf{Evaluation metrics.}
We evaluate the accuracy of the intrinsic parameters and the readout time ratio using the absolute error (AE). Specifically, for the camera intrinsic parameters and the readout time ratio, $\theta \in \{f_x, f_y, c_x, c_y, \gamma\}$, the AE is defined as $\text{AE}(\hat{\theta}) = |\hat{\theta} - \theta|$, where $\theta$ and $\hat{\theta}$ denote the true and estimated values, respectively. In addition, we report the absolute trajectory error (ATE) after alignment to the ground-truth trajectory using a $\mathrm{Sim}(3)$ similarity transformation.

\begin{figure}[t]
\centering
\includegraphics[width=1\columnwidth]{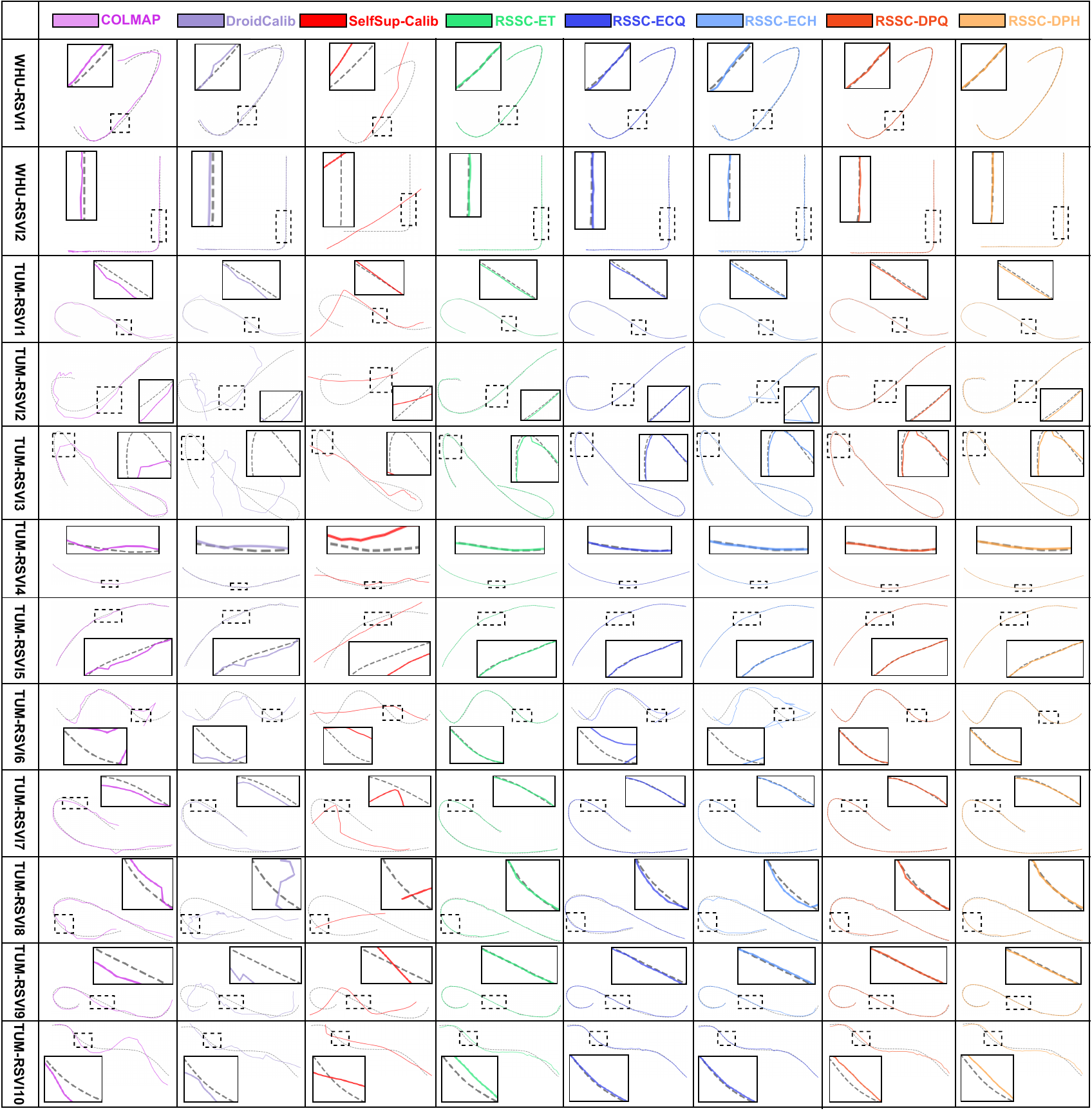}
\caption{Trajectory comparison on the WHU-RSVI~\cite{cao2020whu} and TUM-RSVI~\cite{dai2016rolling} datasets. Each row denotes a different trajectory sequence, while each column corresponds to a specific method. Selected local regions (highlighted by dashed boxes) are magnified and displayed within solid boxes for detailed inspection.}
\label{fig: real_data_result_all}
\end{figure}

\noindent\textbf{Experiment results.}  
Table~\ref{table:self-calibration-results1} presents the intrinsic parameters and readout time ratio AEs, and ATE computed over all sequences in the WHU-RSVI dataset~\cite{cao2020whu}. Tables~\ref{table:self-calibration-results2} and Tables~\ref{table:self-calibration-results3} present the corresponding results for the TUM-RSVI dataset~\cite{dai2016rolling}. Fig.~\ref{fig: real_data_result_all} shows a comparison between the camera trajectories estimated by different methods and the ground truth. Across all metrics, our self-calibration methods consistently outperform the baselines, with RSSC-TE and RSSC-DPQ achieving the best performance. These results demonstrate the accuracy, robustness, and overall effectiveness of the proposed RS self-calibration approach.

\begin{table*}[t]
\centering
\caption{Comparison with RS-aware approache.}
\scalebox{1}[1]{
\begin{tabular}{l l c c c c c c}
\Xhline{1pt}
Dataset & Method & AE($f_x$)& AE($f_y$) & AE($c_x$) & AE($c_y$)  & AE($\gamma$) \\
\Xhline{1pt}

\multirow{6}{*}{TUM-seq1}
& RSCC~\cite{oth2013rolling}   & \textbf{1.14} &	\textbf{0.37} &	\textbf{0.74} &	\textbf{5.97} & \textbf{0.0882}  \\
& RSSC-TE    &  2.53 &	3.49 &	2.37 &	10.26 &	0.1449 
   \\
& RSSC-CEQ   & 1.42 &	1.74 &	1.73 &	8.66 &	0.1130 
   \\
& RSSC-CEH    & 1.15 &	2.50 &	1.74 &	8.87 	&0.1171 
   \\
& RSSC-DPQ   & 1.41 	&1.44 &	2.13 &	8.89 &	0.1273 
   \\
& RSSC-DPH   & 3.17 &	4.33 &	2.46 &	10.77 &	0.1512 
   \\

\midrule

\multirow{6}{*}{TUM-seq2}
& RSCC~\cite{oth2013rolling}   & 3.71 	&0.92 &	\textbf{0.04} &	8.24 &	\textbf{0.0870} 
 \\
& RSSC-TE      & 3.97 &	\textbf{0.01} &	2.21 	&12.90 &	0.1422 
   \\
& RSSC-CEQ    & \textbf{3.04} &	0.23 &	2.45 	& \textbf{6.98} &	0.1021 
    \\
& RSSC-CEH    & 4.19 &	3.07 &	2.93 &	8.26 &	0.1210 
   \\
& RSSC-DPQ   & \textbf{3.04} &	0.64 	&2.11 &	11.25 &	0.1291 
   \\
& RSSC-DPH   & 4.74 	&1.06 &	2.40 &	13.78 &	0.1511 
   \\

\Xhline{1pt}
\end{tabular}
}
\label{table:Comparison with RS-aware approache}
\end{table*}

\noindent\textbf{Comparison with RS-aware approaches.} 
We further include a comparison with RSCC~\cite{oth2013rolling}, a camera calibration method based on RS images. Unlike our approach, RSCC is not a self-calibration method and relies on a visible calibration pattern, operating within a continuous-frame formulation.

Due to the limited availability of datasets that provide both calibration patterns and accurate ground truth, we select two short sequences from the TUM-RSVI dataset~\cite{dai2016rolling} for evaluation, where both methods are applicable. Table~\ref{table:Comparison with RS-aware approache} summarizes the results. RSCC achieves slightly better overall performance, but the difference is marginal. This suggests that our method remains competitive despite not requiring any calibration target, thereby offering greater applicability across a wider range of scenarios.

\begin{table*}[t]
\centering
\caption{Self-calibration results on the TUM-RSVI dataset~\cite{dai2016rolling}. 
The table reports absolute errors (AE) of the intrinsic parameters and readout time ratio, as well as the absolute trajectory error (ATE), for sequences TUM-RSVI1–TUM-RSVI5.}
\scalebox{1}[1]{
\begin{tabular}{l l c c c c c c}
\Xhline{1pt}
Dataset & Method & AE($f_x$)& AE($f_y$) & AE($c_x$) & AE($c_y$)  & AE($\gamma$) & ATE \\
\Xhline{1pt}

\multirow{8}{*}{TUM-1}
& COLMAP~\cite{schoenberger2016sfm} & 44.90 & 32.89 & 19.58 & 63.23 & --- & 0.0291 \\
& DroidCalib~\cite{hagemann2023deep} & 26.49 & 35.69 & 7.98 & 105.09 & --- & 0.0192 \\
& SelfSup-Calib~\cite{fang2022self} & 377.09 & 358.78 & 91.94 & 10.13 & --- & 0.1115 \\
& RSSC-TE & \textbf{1.83} & \textbf{0.27} & 1.86 & 4.43 & 0.1051 & 0.0058 \\
& RSSC-CEQ & 5.47 & 3.78 & \textbf{0.28} & \textbf{0.41} & \textbf{0.0468} & \underline{0.0050} \\
& RSSC-CEH & \underline{3.66} & \underline{2.27} & 0.90 & 3.40 & 0.0949 & \textbf{0.0049} \\
& RSSC-DPQ & 5.03 & 4.61 & \underline{0.40} & \underline{0.78} & \underline{0.0607} & 0.0054 \\
& RSSC-DPH & 8.07 & 3.94 & 3.65 & 7.70 & 0.1455 & 0.0065 \\

\midrule

\multirow{8}{*}{TUM-2}
& COLMAP~\cite{schoenberger2016sfm} & 37.38 & 12.57 & 14.82 & 23.83 & --- & 0.0683 \\
& DroidCalib~\cite{hagemann2023deep} & 241.18 & 195.38 & 11.00 & 58.87 & --- & 0.2345 \\
& SelfSup-Calib~\cite{fang2022self} & 420.30 & 311.57 & 36.00 & 18.69 & --- & 0.2898 \\
& RSSC-TE & 10.47 & \underline{0.65} & 1.38 & 6.50 & 0.1218 & 0.0120 \\
& RSSC-CEQ & \underline{7.02} & 0.68 & 1.56 & \textbf{1.91} & \textbf{0.0842} & \textbf{0.0068} \\
& RSSC-CEH & \textbf{6.67} & 3.48 & \underline{0.87} & 5.63 & 0.1186 & 0.0111 \\
& RSSC-DPQ & 8.77 & \textbf{0.23} & 2.14 & \underline{4.29} & \underline{0.0987} & \underline{0.0101} \\
& RSSC-DPH & 11.30 & 1.18 & \textbf{0.71} & 7.62 & 0.1348 & 0.0130 \\

\midrule

\multirow{8}{*}{TUM-3}
& COLMAP~\cite{schoenberger2016sfm} & 49.03 & 6.30 & 12.83 & 52.50 & --- & 0.0648 \\
& DroidCalib~\cite{hagemann2023deep} & 117.43 & 99.43 & 64.16 & 8.07 & --- & 0.1983 \\
& SelfSup-Calib~\cite{fang2022self} & 117.43 & 99.43 & 64.16 & 8.07 & --- & 0.1163 \\
& RSSC-TE & \underline{0.82} & \underline{1.40} & 0.57 & 1.44 & 0.0804 & 0.0045 \\
& RSSC-CEQ & 2.69 & 2.51 & \textbf{0.26} & 1.55 & \textbf{0.0614} & \textbf{0.0036} \\
& RSSC-CEH & 2.63 & 2.17 & \underline{0.48} & \textbf{1.10} & \underline{0.0661} & \underline{0.0040} \\
& RSSC-DPQ & 1.40 & 2.13 & 0.49 & \underline{1.42} & 0.0717 & 0.0044 \\
& RSSC-DPH & \textbf{0.50} & \textbf{0.77} & 0.50 & 1.51 & 0.0852 & 0.0045 \\

\midrule

\multirow{8}{*}{TUM-4}
& COLMAP~\cite{schoenberger2016sfm} & 13.39 & 13.17 & 4.49 & 2.80 & --- & 0.0085 \\
& DroidCalib~\cite{hagemann2023deep} & 18.68 & 20.59 & 10.47 & 4.04 & --- & 0.0106 \\
& SelfSup-Calib~\cite{fang2022self} & 336.25 & 301.03 & 26.47 & 1.79 & --- & 0.0487 \\
& RSSC-TE & 2.93 & \textbf{0.07} & \underline{3.37} & 0.79 & 0.1261 & \textbf{0.0032} \\
& RSSC-CEQ & \textbf{1.95} & 2.82 & 3.47 & 0.60 & \textbf{0.0018} & \textbf{0.0032} \\
& RSSC-CEH & 3.76 & 3.14 & 3.63 & \textbf{0.15} & 0.0753 & 0.0079 \\
& RSSC-DPQ & \underline{2.39} & \underline{2.00} & \textbf{3.05} & \underline{0.58} & \underline{0.0640} & \underline{0.0034} \\
& RSSC-DPH & 3.46 & 2.18 & 3.81 & 1.14 & 0.1917 & 0.0035 \\

\midrule

\multirow{8}{*}{TUM-5}
& COLMAP~\cite{schoenberger2016sfm} & 13.02 & 27.79 & 17.19 & 12.65 & --- & 0.0155 \\
& DroidCalib~\cite{hagemann2023deep} & 16.29 & 8.39 & 2.14 & 12.71 & --- & 0.0157 \\
& SelfSup-Calib~\cite{fang2022self} & 291.37 & 262.49 & 243.13 & 42.64 & --- & 0.0792 \\
& RSSC-TE & 3.39 & 1.56 & \textbf{0.26} & 2.44 & 0.1013 & 0.0033 \\
& RSSC-CEQ & 3.06 & \textbf{0.50} & 0.78 & \underline{1.91} & \textbf{0.0012} & \underline{0.0031} \\
& RSSC-CEH & \underline{2.58} & 1.50 & \underline{0.64} & \textbf{1.26} & 0.0808 & \textbf{0.0030} \\
& RSSC-DPQ & \textbf{2.41} & \underline{0.93} & 0.84 & 2.28 & \underline{0.0314} & 0.0042 \\
& RSSC-DPH & 4.71 & 5.47 & 1.72 & 2.66 & 0.1792 & 0.0042 \\

\Xhline{1pt}
\end{tabular}
}
\label{table:self-calibration-results2}
\end{table*}

\begin{table*}[t]
\centering
\caption{Self-calibration results on the TUM-RSVI dataset~\cite{dai2016rolling}. 
The table reports absolute errors (AE) of the intrinsic parameters and readout time ratio, as well as the absolute trajectory error (ATE), for sequences TUM-RSVI6–TUM-RSVI10.}
\scalebox{1}[1]{
\begin{tabular}{l l c c c c c c}
\Xhline{1pt}
Dataset & Method & AE($f_x$)& AE($f_y$) & AE($c_x$) & AE($c_y$)  & AE($\gamma$) & ATE \\
\Xhline{1pt}

\multirow{8}{*}{TUM-6}
& COLMAP~\cite{schoenberger2016sfm} & 100.60 & 76.92 & 31.73 & 70.90 & --- & 0.1428 \\
& DroidCalib~\cite{hagemann2023deep} & 313.83 & 253.81 & 204.31 & 52.83 & --- & 0.2732 \\
& SelfSup-Calib~\cite{fang2022self} & 367.07 & 382.76 & 55.61 & 20.94 & --- & 0.2622 \\
& RSSC-TE & \underline{0.93} & \underline{1.08} & \underline{3.97} & \underline{0.30} & \underline{0.0875} & \underline{0.0085} \\
& RSSC-CEQ & \textbf{0.03} & 0.72 & 6.27 & \textbf{0.08} & \textbf{0.0696} & 0.1348 \\
& RSSC-CEH & 3.37 & 5.31 & 7.41 & 2.35 & 0.0895 & 0.3363 \\
& RSSC-DPQ & 1.60 & 1.29 & 6.17 & 0.84 & 0.0739 & 0.0092 \\
& RSSC-DPH & \underline{0.16} & \textbf{0.35} & \textbf{3.07} & 1.25 & 0.0956 & \textbf{0.0086} \\

\midrule

\multirow{8}{*}{TUM-7}
& COLMAP~\cite{schoenberger2016sfm} & 29.77 & 16.47 & 23.98 & 7.21 & --- & 0.0368 \\
& DroidCalib~\cite{hagemann2023deep} & 12.70 & 2.88 & 24.07 & 19.02 & --- & 0.0357 \\
& SelfSup-Calib~\cite{fang2022self} & 347.64 & 380.87 & 83.97 & 1.10 & --- & 0.1708 \\
& RSSC-TE & \underline{0.96} & \underline{1.59} & 3.57 & 1.78 & 0.1242 & \underline{0.0056} \\
& RSSC-CEQ & 0.71 & 2.26 & 6.09 & 1.80 & \textbf{0.0604} & 0.0063 \\
& RSSC-CEH & 1.83 & \textbf{0.77} & \underline{2.50} & \underline{0.70} & \underline{0.0761} & 0.0062 \\
& RSSC-DPQ & 1.93 & 2.65 & 4.64 & 1.44 & 0.0995 & \textbf{0.0049} \\
& RSSC-DPH & \textbf{0.05} & 1.06 & \textbf{2.60} & 2.14 & 0.1386 & 0.0061 \\

\midrule

\multirow{8}{*}{TUM-8}
& COLMAP~\cite{schoenberger2016sfm} & 43.72 & 54.43 & 3.08 & 74.75 & --- & 0.0306 \\
& DroidCalib~\cite{hagemann2023deep} & 168.66 & 115.26 & 123.49 & 30.69 & --- & 0.2172 \\
& SelfSup-Calib~\cite{fang2022self} & 391.30 & 347.33 & 28.97 & 1.54 & --- & 0.2969 \\
& RSSC-TE & 9.40 & 7.79 & 1.65 & \underline{0.71} & 0.1187 & \underline{0.0082} \\
& RSSC-CEQ & \textbf{5.83} & 5.18 & 1.86 & 1.28 & \underline{0.0976} & 0.0094 \\
& RSSC-CEH & 8.80 & \underline{7.64} & \underline{1.51} & \textbf{0.02} & 0.0950 & 0.0119 \\
& RSSC-DPQ & \underline{7.29} & \textbf{4.96} & 1.72 & 5.26 & \textbf{0.0888} & \textbf{0.0075} \\
& RSSC-DPH & 11.02 & 9.82 & \underline{1.51} & 2.13 & 0.1361 & 0.0093 \\

\midrule

\multirow{8}{*}{TUM-9}
& COLMAP~\cite{schoenberger2016sfm} & \textbf{0.07} & \textbf{1.97} & 22.50 & 33.17 & --- & 0.0380 \\
& DroidCalib~\cite{hagemann2023deep} & 142.94 & 138.01 & 87.02 & 57.70 & --- & 0.1269 \\
& SelfSup-Calib~\cite{fang2022self} & 359.77 & 371.34 & 43.22 & 19.97 & --- & 0.1701 \\
& RSSC-TE & \underline{3.74} & 4.31 & 3.74 & \underline{0.74} & 0.1272 & \underline{0.0068} \\
& RSSC-CEQ & 4.51 & \underline{4.11} & \textbf{2.91} & 1.22 & \underline{0.1009} & 0.0075 \\
& RSSC-CEH & 4.19 & 4.23 & 3.80 & 1.78 & \textbf{0.0951} & 0.0083 \\
& RSSC-DPQ & 3.79 & 4.28 & \underline{3.21} & \textbf{0.32} & 0.1068 & \textbf{0.0056} \\
& RSSC-DPH & 3.63 & 4.19 & 4.51 & 2.44 & 0.1553 & 0.0079 \\

\midrule

\multirow{8}{*}{TUM-10}
& COLMAP~\cite{schoenberger2016sfm} & \underline{5.30} & \textbf{6.70} & 7.40 & 2.87 & --- & 0.0620 \\
& DroidCalib~\cite{hagemann2023deep} & \textbf{2.75} & 24.85 & 9.05 & 72.47 & --- & 0.0657 \\
& SelfSup-Calib~\cite{fang2022self} & 359.77 & 329.01 & 78.98 & 9.57 & --- & 0.0721 \\
& RSSC-TE & 7.60 & 8.00 & 0.86 & 0.75 & 0.1100 & 0.0686 \\
& RSSC-CEQ & 8.40 & 7.61 & \textbf{0.76} & \textbf{0.02} & \textbf{0.0597} & \textbf{0.0079} \\
& RSSC-CEH & 5.58 & \underline{7.16} & 1.28 & 3.67 & \underline{0.0780} & \underline{0.0418} \\
& RSSC-DPQ & 7.83 & 8.07 & 1.01 & 0.90 & 0.1004 & 0.0693 \\
& RSSC-DPH & 7.28 & 7.80 & \underline{0.84} & \underline{0.68} & 0.1144 & 0.0683 \\

\Xhline{1pt}
\end{tabular}
}
\label{table:self-calibration-results3}
\end{table*}

\par\vfill\par


%
%
\bibliographystyle{splncs04}
\bibliography{main}